\documentclass[5p]{elsarticle} 
\usepackage{extra-stuff}
\newcommand{\stable}[0]{\mathit{stable}}
\newcommand{\becomesstable}[0]{\mathit{stableup}}
\newcommand{\up}[0]{\mathit{up}}
\newcommand{\down}[0]{\mathit{down}}
\newcommand{\ite}[0]{\mathit{ite}}
\newcommand{\On}[0]{\mathit{On}}

\newcommand{\MinAgg}[0]{\mathit{Min}}
\newcommand{\MaxAgg}[0]{\mathit{Max}}
\newcommand{\stepSize}[0]{\mathit{step}}

\usepackage{fastlr-newcommands}
\begin{document}

\begin{frontmatter}
\title{Reinforcement Learning with Formal Performance Metrics for Quadcopter Attitude Control under Non-nominal Contexts}

\author[1]{Nicola Bernini} 
\ead{nicola.bernini@gmail.com}
\author[1]{Mikhail Bessa} 
\ead{mikhail.bsa@gmail.com}
\author[1]{Rémi Delmas} 
\ead{remi.delmas.3000@gmail.com}
\author[1]{Arthur Gold} 
\ead{arthur.gold.ag@gmail.com}
\author[2]{Eric Goubault\corref{cor1}} 
\ead{goubault@lix.polytechnique.fr}
\author[1]{Romain Pennec} 
\ead{romain.pennec@gmail.com}
\author[2]{Sylvie Putot} 
\ead{putot@lix.polytechnique.fr}
\author[1]{François Sillion} 
\ead{francois.sillion@gmail.com}

\cortext[cor1]{Corresponding author}

\affiliation[1]{organization={Uber ATCP}, 
city={Paris}, country={France}}
\affiliation[2]{organization={LIX, Ecole polytechnique, CNRS, IP-Paris}, 
city={Palaiseau}, country={France}}







\begin{abstract}
We explore  the reinforcement learning approach to designing controllers by extensively discussing  the case of a quadcopter attitude controller. We provide all details allowing to reproduce our approach, starting with a model of the dynamics of a {\it crazyflie 2.0} under various nominal and non-nominal conditions, including partial motor failures and wind gusts. 
We develop a robust form of a signal temporal logic to quantitatively evaluate the vehicle's behavior and measure the performance of controllers. 
The paper thoroughly describes the choices in training algorithms, neural net architecture, hyperparameters, observation space in view of the different performance metrics we have introduced. We discuss the robustness of the obtained controllers, both to partial loss of power for one rotor and to wind gusts and finish by drawing conclusions on practical controller design by reinforcement learning.  
\end{abstract}


\end{frontmatter}

\section{Introduction}


Neural net based control is now widely used in control. For instance, reinforcement learning is known to be linked to optimal control \cite{bertsekas2019reinforcement}. Very impressive real-life experiments have shown how practical reinforcement learning and privileged learning can be 
\cite{deepdroneacrobatics}, but have somehow masked the enormous amount of experiments and heuristics that had to be learned in the process. 
Indeed, we are still in need for a full understanding of what advantages and performances we can gain from learning-based control, and what level of formal guarantees we can reach, either at design or at verification time.

This paper extends our HSCC 2021 article~\cite{hscc2021} with a more complete description of several aspects including the modeling, lessons that have been learned, and most importantly the description of the logic that has been used for evaluating performances of our neural net controllers, as well as new results concerning some spurious correlations that appeared in all attitude controllers that we trained. 

We concentrate here on low-level controls, and more specifically attitude control for  quadcopters. These controllers have the advantage of being understandable - performances being easily measurable -, well studied in the literature, and essential to all higher-level controls and path tracking algorithms. We focus on reinforcement learning (RL) methods, which are close to control and more particularly optimal control. Furthermore,  RL has experienced tremendous progress over the past few years, with modern continuous state and action spaces training algorithms such as Soft Actor Critic (SAC) \cite{SAC} and Twin-Delayed Deep Deterministic Policy Gradients (TD3) \cite{TD3}. 

A common belief is that learning-based control would be more robust to perturbations than e.g. PIDs, or at least could be trained to be more robust. Indeed, even a rather small neural net can encode a much more complex feedback control function than a simple PID, but this is commonly believed to be at the expense of formal guarantees. Also, the current zoology of training methods and architecture choices makes it difficult to fully understand the range of possible results. 

This paper studies some of these aspects on the fundamental case of an attitude controller for  the crazyflie 2.0 \cite{nanoquadcop} quadcopter. We first present in \Cref{sec:model-control-quad} a non-linear ODE model for simulating the dynamics of a quadcopter, and extend it to account for partial motor failures, aerodynamic effects and wind gusts. We then present a flexible training platform with various neural net architectures and algorithms in  
\Cref{sec:training}, discuss performance evaluation using a robust signal temporal logic in \Cref{sec:perf-observers}, and describe our experimental setup in \Cref{sec:experiments}. Finally we discuss experimental results in \Cref{sec:exp-results}. 



This paper develops in detail the following research items:
\begin{enumerate}
    \item we develop a neural-net based control study case, after modeling a quadcopter's dynamics, including aerodynamic effects and partial power loss on motors 
    \item we discuss the effect of the chosen training algorithm, neural net architecture, reduced observable state spaces and hyperparameters on the performance of the controller, and on the RL training process     
    \item we present our experimental platform, which allowed us to compare more than 16,000 
    parameter choices 
    \item we develop Signal Temporal Logic observers to assess controller performance in a precise manner 
    \item we demonstrate high-quality attitude control using RL, for a relevant set of queries 
    \item we show that these controllers have a certain built-in robustness in non-nominal cases, with respect to partial failures of actuators and perturbations such as wind gusts. 
    \item we discuss in details the lessons learned in reinforcement learning, while applying it to the problem of synthesizing quadcopter attitude controllers 
\end{enumerate}


\section{Related work}

This paper is based on, and compared with, the following work: 

\paragraph{RL in control}

Reinforcement learning in control has been advertised, since 
\cite{Sutton}, for the possibility to be more adaptative than classical methods in control such as PIDs. RL's close relationship with optimal control (the reward function is dual to the objective function) also makes it particularly appealing for applications to control, see e.g. \cite{bertsekas2019reinforcement}. 

Recently model-based reinforcement learning has been successfully used to train controllers without any initial knowledge of the dynamics and in a data-efficient way. For instance, in \cite{lambert2019low}, a learning-based model predictive control algorithm has been used to synthesize a low level controller. In \cite{yoo2020hybrid}, a hybrid approach is proposed, combining the model based algorithm PILCO \cite{Deisenroth2011PILCOAM} and a classic controller like a PD or a LQR controller. 



In this paper, we focus on model-free algorithms because of their generality and because we have high fidelity models available for quadrotors, such as the crazyflie 2.0 \cite{nanoquadcop}. More specifically we concentrate on actor-critic learning which has undergone massive improvements over the last few years with DDPG \cite{DDPG}, SAC \cite{SAC}, TD3 \cite{TD3}, and compare it with the popular PPO method \cite{PPO}.

The high dimensionality of the full Markovian observation space is a challenge for training, prompting for a study of different choices for the sets of states observed by RL: we consider sub-spaces of the full Markovian observation space, where we leave out the states which have the least effect on the dynamics of the quadcopter. This is linked to partially observed Markov Decision Processes and Non Markovian learning, see e.g. \cite{NMR}. 

We also study the robustness of our neural nets, as well as the specific training of the neural net controller to be able to handle disturbances (wind gusts, partial motor failures). These issues may be linked to robust MDPs \cite{RMDP}, but we have stuck to the classical (PO)MDP approach here, for which we have a wealth of tools and techniques available.

\paragraph{RL for quadcopters, and attitude control}

Most papers have been focusing on higher-level control loops, with the notable exception of \cite{rl}, which serves as the basis of our work.
We improve the results of \cite{rl} by considering more recent training algorithms (SAC and TD3), finer performance measures, and refined physical models (in particular perturbations due to partial motor failures and wind gusts). 
The closest other works related to attitude control for quadcopter are \cite{simtoreal}, \cite{fei2020learn}, \cite{Koning} and \cite{stockholm}. 

In \cite{simtoreal}, the goal is to stabilize a quadcopter in hover mode, from various initial conditions (including initial angular rates). The authors also consider perturbations to the dynamics, which are more predictable than ours: motor lag and noise on sensors. 
 
In \cite{fei2020learn}, the objective is to control a quadcopter under cyber-attacks targeting its localization sensors (gyroscope and GPS) and motors. The authors consider (partial) motor failure (a limit on its maximal power, just like we do), but not wind gusts. Contrarily to most approaches including ours, their controller combines a classical controller and a neural net.

 
In \cite{Koning} the authors discuss the training of a neural net controller for both attitude and position. They observe that it is difficult to train both aspects at the same time, whereas separating control in hover mode (acting mostly on the attitude) and control in position seems to work better. 
The learning process is based on a full state observation plus the difference with the target state. We extend this work first in discussing the simplification of the observed states, then in more rigorously defining observation metrics for offsets and overshoots.

In \cite{stockholm}, the author considers neural nets for controlling roll, pitch, yaw rate and thrust, which is similar to the problem we are studying here, and attempts to train the controller such that it can accommodate motor and mass uncertainties within given bounds. In contrast, we deal with uncertainties such as wind gusts and motor failures, following known parametric models.




\paragraph{Signal Temporal Logics}
The study of reinforcement learning under temporal logic
specifications has gained a lot of interest in recent years.
In a discrete and finite state setting, in \cite{2015wencorrectbysynthesis} a linear-time temporal logics (LTL) property observer automaton
is composed with the system MDP to allow blocking unsafe actions
during training. In \cite{2019gaoreducedvariance, 2019hasanbeigreinforcement} rewards are modulated depending on
the observer state, and a model-free approach is proposed
in \cite{2019hasanbeigtowards} using Limit Deterministic B\"uchi
Automata. \emph{Shielding} \cite{2018alshiekhshielding} simultaneously trains an
optimal controller and a \emph{shield} that corrects the
LTL-formula violating actions. 
The method requires a fully explicit model
of the environment MDP and builds the product of the orignal MDP with the 
property monitor. Later works extend shielding to the continuous
\cite{2019zhangosafemultiagentrl, 2020bastanisafe} and online
\cite{2019bastanionlineshielding} cases, assuming an embeddable
predictive environment model is available, but only handle simple state invariants.

Temporal logics with quantitative
semantics such as
Metric Interval Temporal Logic (MITL)
\cite{2009fainekosrobustnessMITL}, Signal Temporal Logic (STL)
\cite{2013donzestl} \dots, have been studied in relation with
reinforcement learning. Robust interpretation yields a real number
indicative of the distance to the falsification boundary. STL has
seen numerous extensions improving expressiveness and signal classes
\cite{2014brimstlstar, 2015akazakiaveragemtl, 2019bakhirkinbeyondstl, 2019abbasrobustnessgeneral} as
well as smooth differentiable semantics \cite{2019haghighismoothstl, 2019mehdipouragm, 2021gilpinsmoothstl}.
Solutions to well known dimension and magnitude
mismatch in robust STL interpretation were proposed recently in
\cite{2019zhangbandits} but have not yet been used in a RL
setting.
STL usages are varied:
In
\cite{2016aksarayqlearningrobust}, Q-learning is used to train a policy maximizing both
the probability of satisfaction and the expected robustness of a given STL specification;
The approach requires storing previously
visited states in the MDP in addition to the original MDP state,
yielding a high dimensional system and limiting learning
efficiency. 
In \cite{2019lilogicguidedsaferlbarrier} the authors derive barrier functions
from robust temporal logic specifications, either to modulate
rewards during training 
or to control the switch from an optimal and potentially unsafe
controller to a safe backup controller
\cite{2019lindemanncontrolbarrierstl}.

In summary, existing methods focused on the training phase either suffer
from dimensionality and combinatorial explosion, require expected
robustness approximations, or are strongly tied to the
Q-learning algorithms. 

Considering our goal is to study a large hyper-parameter space for
training controllers and we need to quantify controller performance rigorously, we used an expressive yet tractable variant of
STL \cite{2019bakhirkinbeyondstl} to specify properties and assess trained controllers offline, separately after training. The next steps will be to start
using STL-derived reward signals during training on the most promising
architectures.

\section{Modelling and control of a crazyflie 2 quadrotor}

In this section, we present the dynamical model of the crazyflie quadrotor \cite{goubault_putot,quadcopter_model} and we augment it with partial motor failures and wind gusts modelling.

\label{sec:model-control-quad}
\subsection{Nominal model}
\label{sec:modelling}


\begin{figure}[htbp]
    \centering\noindent
    \begin{subfigure}[t]{.49\linewidth}
        \centering
        \includegraphics[height=3cm]{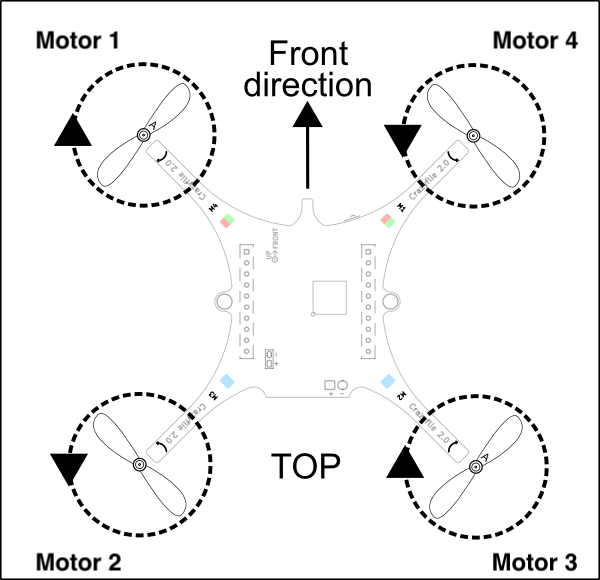}
        \caption{Motors' controls}
        \label{fig:crazyflie-motors}
\end{subfigure}
    \hfill
    \begin{subfigure}[t]{.49\linewidth}
        \centering
        \includegraphics[height=3cm]{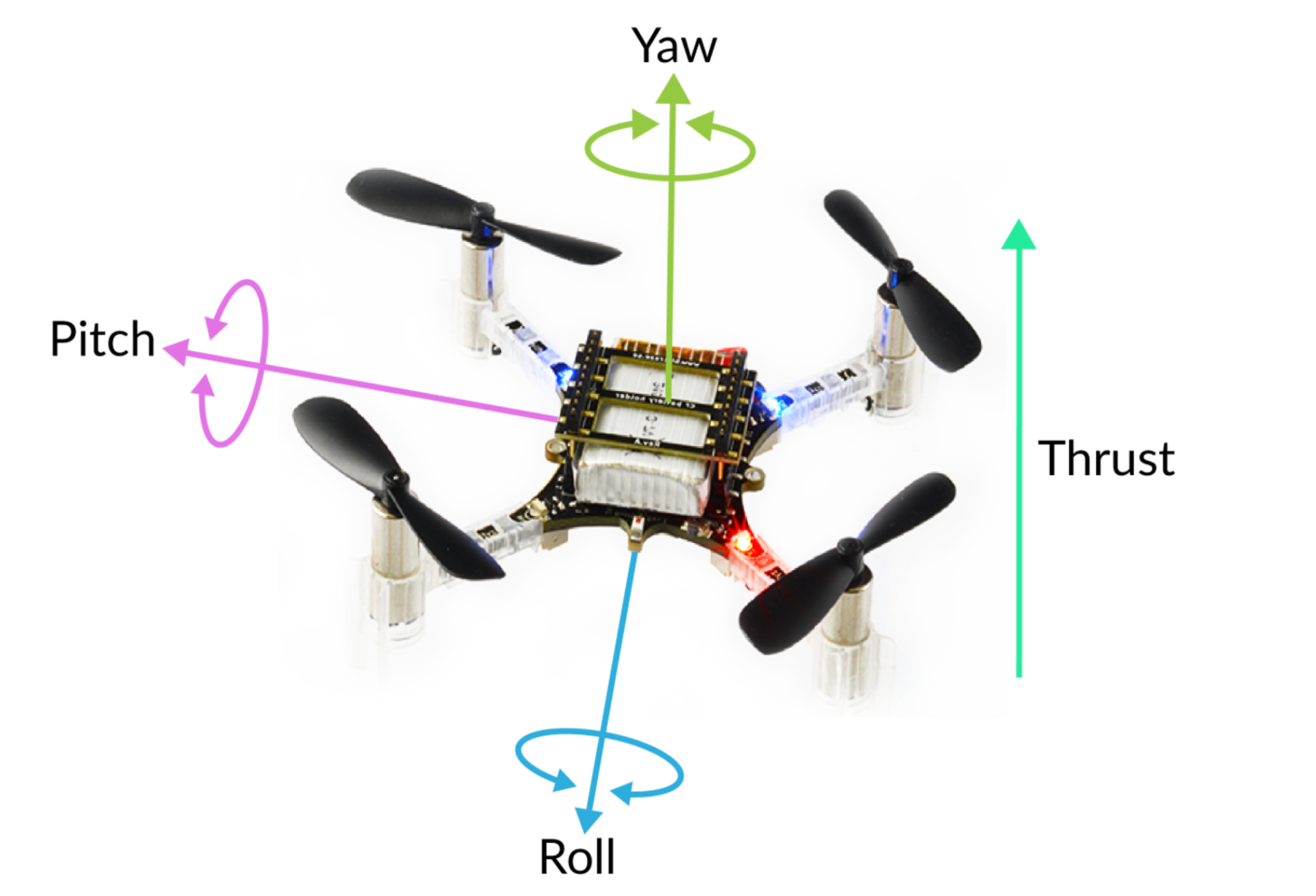}
        \caption{Principal axes}
        \label{fig:crazyflie-axes}
    \end{subfigure}
    \caption{Crazyflie 2.0 -- source: \texttt{http://www.bitcraze.io} \cite{bitcraze} CC BY-SA 3.0}
        \label{fig:crazyflie}
\end{figure}


We study the dynamics on the vertical axis and the pitch rate, roll rate and yaw rate control (4 degrees of freedom), with the following state variables: 
the vertical position in the world frame $z$,
the linear velocity of the center of gravity  
in the body-fixed frame with respect 
to the inertial frame $(u, v, w)$,
the angular orientation represented by the Euler angles $(\phi, \theta, \psi)$ where $\phi$ is the roll angle 
$\theta$ is the pitch angle and $\psi$ is the yaw angle, 
the attitude or angular velocity with respect to the body frame $(p, q, r)$ with $p$ the roll rate, $q$ the pitch rate and $r$ the yaw rate.

The Crazyflie 2.0 linear velocities are controlled through the angular velocities and the angular velocities are controlled through rotor thrust differential. For instance, to increase the pitch rate $q$, $Motor_2$ and $Motor_3$ rotor speeds should be higher than $Motor_1$ and $Motor_4$ (see Figure \ref{fig:crazyflie-motors}). As there is symmetry, it works similarly for the roll rate $p$ (with $Motor_4$ and $Motors_3$ vs. $Motor_1$ and $Motor_2$ instead). However, the yaw rate $r$ is controlled through the gyroscopic effect. To make the quadcopter rotate clockwise in the x-y plane, the rotor speeds of the clockwise rotating motors ($Motor_2$ and $Motor_4$) should be higher than those of the counterclockwise rotating ones ($Motor_1$ and $Motor_3$). \\
Using Newton's equations given a \emph{thrust} force and moments $M_x$, $M_y$ and $M_z$ exerted along the three axes of the quadcopter, 
and using the rotation matrix $R$ from the body frame to the inertial frame, 
\begin{equation*}
R = 
\begin{pmatrix}
c_\psi c_\theta & c_\psi s_\theta s_\phi -c_\phi s_\psi & s_\psi s_\phi +c_\psi c_\phi s_\theta \\
c_\theta s_\psi & c_\psi c_\phi +s_\psi s_\theta s_\phi & c_\phi s_\psi s_\theta -c_\psi s_\phi \\
-s_\theta & c_\theta s_\phi & c_\theta c_\phi
\end{pmatrix}
\end{equation*}
\noindent (and $R^{-1}$ is the transpose of $R$) 
the Translation-Rotation kinematics and dynamics \cite{quadcopter_model} lead to a 10-dimensional non-linear dynamical system:  
\begin{equation}
\label{eq:dynamic}
\left\{
\begin{alignedat}{2}
\dot{z} &= -s_\theta u + c_\theta s_\phi v + c_\theta c_\phi w &\qquad \dot{\theta} &= c_\phi q  - s_\phi r \\
\dot{u} &= rv - qw + s_\theta g & \dot{\psi}   &= \tfrac{c_\phi}{c_\theta} r + \tfrac{s_\phi}{c_\theta} q \\
\dot{v} &= -ru + pw - c_\theta s_\phi g & \dot{p} &= \tfrac{I_y - I_z}{I_x} qr + \tfrac{1}{I_x} M_x \\
\dot{w} &= qu - pv - c_\theta c_\phi g + \tfrac{F}{m} & \dot{q} &= \tfrac{I_z - I_x}{I_y} pr + \tfrac{1}{I_y} M_y \\
\dot{\phi}   &= p + c_\phi t_\theta r  + t_\theta s_\phi q & \dot{r} &= \tfrac{I_x - I_y}{I_z} pq + \tfrac{1}{I_z} M_z \\
\end{alignedat}
\right.
\end{equation} 
\noindent
writing $c_{x}$ as a short for $cos(x)$, $s_{x}$ for $sin(x)$ and $t_{x}$ for $tan(x)$. 
$F$ is the sum of the individual motor thrusts, and $I_x$, $I_y$, $I_z$ are the quadcopter's moments of inertial around the $x$, $y$ and $z$ axes, respectively. 

Instead of controlling directly each rotor speed, the four commands $thrust$, $cmd_{\phi}$, $cmd_{\psi}$ and $cmd_{\theta}$ are used to deduce the PWM (Pulse Width Modulation)values to apply to each motor, \Cref{eq:PWM}: 

\begin{equation}
\label{eq:PWM}
PWM = 
\begin{bmatrix}
  {PWM}_1 \\
  {PWM}_2 \\
  {PWM}_3 \\
  {PWM}_4
\end{bmatrix} = 
\begin{bmatrix}
  1 & -\nicefrac{1}{2} & -\nicefrac{1}{2} & -1\\
  1 & -\nicefrac{1}{2} & \phantom{-}\nicefrac{1}{2} & \phantom{-}1\\
  1 & \phantom{-}\nicefrac{1}{2} & \phantom{-}\nicefrac{1}{2} & -1\\
  1 & \phantom{-}\nicefrac{1}{2} & -\nicefrac{1}{2} & \phantom{-}1\\
\end{bmatrix}
\begin{bmatrix}
thrust \\
cmd_\phi \\
cmd_\theta \\
cmd_\psi \\
\end{bmatrix}
\end{equation} \\

PWMs are linked to rotation rates $\Omega$: 
$   \Omega = 
    [\omega_1 \ 
    \omega_2 \ 
    \omega_3 \ 
    \omega_4 
]^\top
 = C_1 PWM + C_2$. 
Finally, we deduce the input force and moments from the squared rotation rates,  \Cref{eq:dynamic}, with force and momentum equations $[F \ M_x \ M_y \ M_z]^\top$ equal to:

\begin{equation} \label{eq:fulldynamic}
\begin{bmatrix}
  C_T \big(C_1^2(cmd_{\theta}^2 + cmd_{\phi}^2 + 4cmd_{\psi}^2 + 4thrust^2) \\
  \mbox{ } \ \ \ + 8C_1 C_2 thrust + 4C_2^2 \big)\\
  4C_Td\big(C_1^2 (cmd_{\phi}thrust - cmd_{\theta}cmd_{\psi}) +  C_1C_2cmd_{\phi} \big)  \\
  4C_Td\big(C_1^2 (cmd_{\theta}thrust - cmd_{\phi}cmd_{\psi}) +  C_1C_2cmd_{\theta}  \big)  \\
  2C_D\big(C_1^2(4cmd_{\psi}thrust - cmd_{\phi}cmd_{\theta}) + 4C_1C_2cmd_{\psi} \big)\\
\end{bmatrix} \\
\end{equation}
\noindent 
The physical and constant parameters we are using for the crazyflie are obtained by merging data from \cite{quadcopter_model} and \cite{nanoquadcop} and listed in Table~\ref{params2}: 

\begin{table*}[h]
\centering
\begin{tabular}{@{}llrl@{}}
    \toprule
    Param & Description & Value & Unit \\
    \midrule
    $I_x$ & Inertia about x-axis & \num{1.657171e-5}  & \si{\kilo\gram\square\metre} \\
    $I_y$ & Inertia about y-axis & \num{1.6655602e-5} & \si{\kilo\gram\square\metre} \\
    $I_z$ & Inertia about z-axis & \num{2.9261652e-5} & \si{\kilo\gram\square\metre} \\
    $m$ & Mass & \num{0.028} & \si{\kilo\gram} \\
    $g$ & Gravity & \num{9.81} & \si{\metre\per\square\second} \\
    $C_T$ & Thrust Coefficient & \num{1.285e-8}  & \si{\newton\per\square\radian\square\second} \\
    $C_D$ & Torque Coefficient & \num{7.645e-11} & \si{\newton\per\square\radian\square\second} \\
    $C_1$ & PWM to $\Omega$ factor & \num{0.04076521} & - \\
    $C_2$ & PWM to $\Omega$ bias   & \num{380.8359}   & - \\
    $h$ & z rotor wrt CoG & 0.005 & \si{\meter} \\
    $d$ & Arm length & $0.046/\sqrt{2}$ & \si{\meter} \\
    $p_{max}$ & Maximum motor PWM & \num{65535} & - \\
    \bottomrule
\end{tabular}
\caption{Parameters for the crazyflie 2.0 model}
\label{params2}
\end{table*}



\subsection{Motor failure}

\label{sec:motorfailure}

We suppose that the quadcopter may experience a power loss on motor 1. 
This partial failure is modeled as a saturation of the maximum PWM, with a factor between 0.8 and 1.


 
Since quadcopter controls rely on differential thrust between motors, motor failures are very difficult to cope with. In order to keep a constant yaw when one motor is failing, the gyroscopic effect must be made equal to zero, for instance by having the two motors rotating in the opposite direction match the saturation of the faulty motor. The same idea applies to pitch and roll axes. 

Therefore, if the failure is not too harsh, and the target states are not too demanding, it is {\it a priori} feasible to recover some control of the faulty quadrotor by saturating all four motors in the same way. 

In this paper, we will look at two potential solutions to control in the presence of partial motor failure. The first one is to look at how robust a controller that has been designed for nominal cases (i.e. without partial motor failures) is. The other one is to train, using reinforcement learning, a controller optimized for a variety of non-nominal situations. 





\subsection{Wind gusts}

\subsubsection{Aerodynamic effects}
\label{sec:aero}

In \Cref{eq:dynamic}, we neglected all aerodynamic effects. 
When we take into account aerodynamic forces, an extra force $F^a$ is exerted on the quadcopter that depends on the wind speed and direction relative to the quadcopter, the angular velocities of the rotors and extra moments $M^a_x$, $M^a_y$ and $M^a_z$. We follow the full aerodynamic model of \cite{nanoquadcop} with the coefficients measured for a crazyflie 2.0, where the effect of the wind on the structure is neglected with respect to the effect on the rotors, and the blade flipping effect (due to elasticity of the rotor) is also neglected. 

The extra force $F^a$ can be decomposed as the sum of the four extra aerodynamic forces on rotor $i$ ($i=1,\ldots,4$), that can be modelled as depending linearly on the rotors angular velocities, and linearly on the wind relative speed with respect to rotors. Other models \cite{Bangura} include blade flipping and other drag effects, but the induced drag we are modelling is the most important one for small quadrotors with rigid blades. 
We use $f^i= \Omega_i K W^r_i$  for the aerodynamic force exerted on rotor $i$
in the inertial frame, where $K$ is the drag coefficients matrix, 
$W^r_i$ is the relative wind speed as seen from rotor $i$, in the body frame, i.e. $W^r_i=(u_i, v_i, w_i)-R^T W_a$ with $W_a$ the absolute wind speed in the inertial frame, $(u_i, v_i, w_i)$ being the linear velocities of the rotors in the body frame, $R$ is the rotation matrix from the body frame to the inertial frame ($R^T$ is its inverse, i.e. its transpose here), and $\Omega_i $ is the absolute value of the  angular velocity of the $i$-th rotor. 

The drag coefficients we are using for the crazyflie are one of the models of \cite{nanoquadcop}:
\begin{equation*}
K =
\begin{pmatrix}
-9.1785 & 0 & 0 \\
0 & -9.1785 & 0 \\
0 & 0 & -10.311
\end{pmatrix}
10^{-7} kg.rad^{-1}
\end{equation*}

For the crazyflie, $\Omega_i=C_1 PWM_i+C_2$, where the expression $PWM_i$ depends on $thrust$, $cmd_{\phi}$, $cmd_{\theta}$ and $cmd_{\psi}$ as given by \Cref{eq:PWM}.

The linear velocities of rotors can be computed as follows:
\begin{equation*}
\begin{aligned}
\pmat{u_j \\ v_j \\ w_j} &= \pmat{p \\ q \\ r}
  \times \pmat{d c_j \\ d s_j \\ h} + \pmat{u \\ v \\ w} 
  &=
  \begin{pmatrix}
    \phantom{-}qh-rds_j+u \\
    -ph+rdc_j+v \\
    pds_j-qdc_j+w
  \end{pmatrix}
\end{aligned}
\end{equation*}
\noindent $(u,v,w)$ are the linear velocities of the
center of mass of the quadrotor in the body frame, $(p,q,r)$ are the angular velocities of the quadrotor (see Section \ref{sec:modelling}).
$d$ is the length of the arm linking the center of the drone to any of the four motors, and for $j \in \{1, 2, 3, 4\}$, 
$c_j=sin\big(\frac{\pi}{2}(j-1)+\frac{3\pi}{4}\big)$ and $s_j=cos\big(\frac{\pi}{2}(j-1)+\frac{3\pi}{4}\big)$ are such that  $(c_j,s_j,h)$ is the coordinate of rotor $j$ in the body frame, with the center of mass being the origin.

Now, we add to the second term of \Cref{eq:dynamic} for $\dot{u}$, $\dot{v}$, $\dot{w}$ the aerodynamic force $F^a=(F^a_x,F^a_y,F^a_z)$ divided by $m$, and to moments of \Cref{eq:fulldynamic}, the aerodynamic moments $M^a=(M^a_x, M^a_y,$ $M^a_z)$ with 
$F^a  = f_1+f_2+f_3+f_4$
and $M^a = (dc_1,ds_1,h)\wedge f_1 + (dc_2,ds_2,h) \wedge f_2
      + (dc_3,ds_3,h)\wedge f_3+(dc_4,ds_4,h) \wedge f_4$.

We derive the full dynamics of the quadcopter considering aerodynamic effects, and only write below the modified equations: 
\begin{equation}
\label{eq:aerodynamic}
\left\{
\begin{alignedat}{2}
\dot{u} &= rv - qw + s_\theta g + \tfrac{F^a_x}{m} \\
\dot{v} &= -ru + pw - c_\theta s_\phi g\!+\!\tfrac{F^a_y}{m} \\
\dot{w} &= qu - pv - c_\theta c_\phi g\!+\!\tfrac{F+F^a_z}{m} \\
\dot{p} &= \tfrac{I_y - I_z}{I_x} qr + \tfrac{1}{I_x} (M_x+M^a_x) \\
\dot{q} &= \tfrac{I_z - I_x}{I_y} pr + \tfrac{1}{I_y} (M_y+M^a_y) \\
\dot{r} &= \tfrac{I_x - I_y}{I_z} pq + \tfrac{1}{I_z} (M_z+M^a_z) \\
\end{alignedat}
\right.
\end{equation} 

\subsubsection{Wind models}

There are two main types of models in the literature, represented by e.g. Discrete Wind Gust and von K\'arm\'an Gust or Dryden Gust models. Von K\'arm\'an gusts and Dryden gusts are stochastic gust models (homogeneous and stationary gaussian processes) characterized by their power spectral densities for the wind's three components, Dryden gusts being an approximation of Von K\'arm\'an gusts. 

The Discrete Wind Gusts model consists in a explicit and deterministic representation of wind gusts as half period cosine perturbations (\cite{demourant1}, eq. (45)).
We focus on this model because it is widely used for aircraft certification (using dozens of discrete wind gusts with different magnitudes and scales). 

A discrete wind gust is characterized by its fixed direction, magnitude and scale, and lasts for a half period during which wind speed increases until it reaches its maximum intensity. The absolute wind velocity is given as a function of time as, using the same notations as in \Cref{sec:aero}:
$W_a(t) =
  \frac{A_g}{2} \left(1 - cos\big(\frac{\pi (t-t_0)}{\delta}\big)\right) V_d$ if $  t_0 \leq t \leq t_0+2\delta$, 0 otherwise, 
where $A_g$ is the maximal magnitude of the wind gust, $\delta$ is the half life of the gust, and $V_d$ is a normalized vector in $R^3$, which is the wind (absolute) direction. 
        
 
\subsection{PID Control}

As in \cite{rl}, the objective is to train only the attitude controller, and not the altitude one. We therefore use a PID for controlling $z$. We will also need some idea of what a standard PID controller may achieve in terms of performance, and robustness to wind gusts and failures. For this, we will primarily use one of the altitude and attitude PID controller implemented in the crazyflie 2.0. 
Given setpoints $z_{sp}$, $p_{sp}$, $q_{sp}$ and $r_{sp}$, the quadrotor is controlled using a PID controller (called PID1 in the sequel) which is the one of \cite{goubault_putot}: 
\begin{equation}
\label{eq:control1}
\left\{
\begin{aligned}
thrust &= 1000 \big(25(2(z_{sp} - z) - w) \\ & \qquad + 15 \smallint (2(z_{sp} - z) - w)\dt\big) + 36000 \\
cmd_{\phi} &= 250 (p_{sp} - p) +  500 \smallint (p_{sp} - p) \dt \\
cmd_{\theta} &= 250 (q_{sp} - q) +  500 \smallint (q_{sp} - q) \dt \\
cmd_{\psi} &= 120 (r_{sp} - r) + 16.7 \smallint (r_{sp} - r) \dt
\end{aligned}
\right.
\end{equation}

But as we will see, the attitude controller implemented in the crazyflie 2.0 is not very reactive, most probably for ensuring that the altitude is very securely controllable (since too much reactivity in pitch and roll means sudden loss of vertical speed). In order to give an idea of what we could observe as best performance, we 
also designed a specific PID for attitude, that we call PID2, which is much more reactive:  

\begin{equation}
\label{eq:control2}
\left\{
\begin{aligned}
thrust &= 3000(z_{sp} - z) \\ & \qquad + 300 \smallint (z_{sp} - z) \dt - 500 \dot{z} + 48500 \\
cmd_{\phi} &= 1000 (p_{sp} - p) +  400 \smallint (p_{sp} - p) \dt - 40 \dot{p} \\
cmd_{\theta} &= 1000 (q_{sp} - q) +  400 \smallint (q_{sp} - q) \dt - 40 \dot{q}\\
cmd_{\psi} &= 2000 (r_{sp} - r) + 1000 \smallint (r_{sp} - r) \dt - 100 \dot{r}
\end{aligned}
\right.
\end{equation}

\section{Training}

\label{sec:training}

\subsection{Underlying Markov decision process}

\label{sec:Markov}

Reinforcement learning is designed to solve Markov decision problems. At each discrete time step $k=1, 2, \ldots$, the controller observes the state $x_k$ of the Markov process, selects action $a_k$, receives a reward $r_k$, and observes
the next state $x_{k+1}$. As we are dealing with Markov processes, the probability distributions for $r_k$ and $x_{k+1}$ depend only on $x_k$ and $a_k$. Reinforcement learning tries to find a control policy, i.e. a mapping from states to actions, in the form of a neural net, that maximizes at each time step the expected discounted sum of future reward. 

For the attitude control problem at hand, the set of Markovian states is $thrust$, $p$, $q$, $r$, $err_p=p_{sp}-p$, $err_q=q_{sp}-q$, $err_r=r_{sp}-r$ (where $(p_{sp}, q_{sp}, r_{sp})$ is the target state, or "plateau" we want to reach), in the nominal case (similarly to what is done in e.g. \cite{Koning}). We will also consider partially observed Markov processes, with only subsets of states for improving sampling over smaller dimensional states, by leaving out those states which should have less influence on the dynamics: our first candidate is to leave out thrust, which appears only as second order terms in the moments calculation, \Cref{eq:fulldynamic}, and also, $p$, $q$, $r$ that are second order in the formulation of the angular rates, again in \Cref{eq:fulldynamic}. 
We do not consider here adding past information, classical in non Markovian environments \cite{NMR}, that has been used for attitude control in e.g. \cite{rl}, but increases the dimension by a large amount. 

In the case of partial motor failure, we add the knowledge of the maximum thrust for faulty motor 1, as a continuous variable between 80\% and 100\%. In the case of aerodynamic effect and wind gusts, we add the knowledge of the maximal magnitude and direction (in the inertial frame) of the incoming wind. In both cases, it can effectively be argued that it is possible to detect failures in almost real time, and to measure (or be given from ground stations) maximum winds and corresponding directions, in almost real time as well. In the case of wind-gusts, Markovian states include also the linear velocities $u$, $v$ and $w$, since wind gusts are only defined in the inertial frame, and the induced aerodynamic effects depend on relative wind speed. 

With a view to solving optimal control problems (or Model-Predictive like control), we choose to use a reward function 
which is a measure of the distance between the current attitude $(p,q,r)$ with $(p_{sp}, q_{sp}, r_{sp})$, the target attitude (similar to the one used in \cite{rl}): 
$$r(s) = - min\left(1, \frac{1}{3\Omega_{max}} \left\lVert \Omega^* - \Omega \right\rVert\right) $$ 
\noindent $\Omega_{max}$ being the maximal angular rate that we want to reach for the quadcopter, and $\Omega$ is the angular rate vector $(p,q,r)$ which is part of the full state $s$ of the quadcopter.


\subsection{Neural net architecture}

\label{sec:neuralnetarchi}


Neural nets, such as multiple layer perceptrons (MLP) with RELU activation, can efficiently encode all piecewise-affine functions \cite{RELUnet}. It is also known \cite{MPC} that the solution to a quadratic optimal control (MPC) problem for linear-time invariant system is piecewise-affine. Furthermore, there are good indications that this applies more generally, in particular for non-linear systems \cite{nonlinearmpc}. This naturally leads to thinking that MLPs with RELU networks are the prime candidates for controlling the attitude with distance to the objective as cost (or reward). In some ways, the resulting piecewise-affine function encodes various proportional gains that should be best adapted to different subdomains of states, so as to reach an optimal cumulated (and discounted, here with discount rate $\gamma=0.99$\footnote{All other parameters, learning rates in particular are the standard ones of Stable Baselines 2.7.0}) distance to the target angular rates, until the end of training.

In theory \cite{Ferlez}, one could find a good indication of the architecture of the neural net in such situations, but the bounds  that are derived in \cite{Ferlez} are not convenient for such a highly complex system. It is by no means obvious what architecture will behave best, both for training and for actual controller performance, although a few authors argue that deeper networks should be better, see e.g. 
\cite{LUCIA2018511}. 

Architectures that have been reported in the literature for similar problems are generally alike. 
In \cite{simtoreal}, 
the neural net is a Multi-Layer Perceptron (MLP) with two layers of 64 neurons each, and with $\mathrm{tanh}$ activation function. 
In \cite{fei2020learn}, the part of  the controller which is a neural net is a MLP with two layers of 96 neurons each and $\mathrm{tanh}$ activation function, 
whose input states (observation space) are all states plus the control. 
In \cite{Koning}, the hover mode neural net controller, which is the most comparable to our work, is a MLP with two layers of 400 and 300 neurons respectively, with RELU activation for hidden layers and $\mathrm{tanh}$ for the last layer. 
In \cite{stockholm}, the resulting architecture is a two layers MLP with 128 neurons on each layer, and RELU activation function. 



We will report experiments with one to four layers, and with 4, 8, 16, 32 or 64 neurons per layer, with RELU activation function (except for the rescaling of the output, using $\mathrm{tanh}$). We limit the reporting of our experiments to these values since we observed that these were enough to find best (and worst) behaviours. 

\subsection{Training algorithms}

\label{subsec:training}

The first three algorithms we are discussing in \Cref{sec:experiments}, DDPG \cite{DDPG}, SAC \cite{SAC} and TD3 \cite{TD3} are all off-policy, actor-critic methods, which are generally considered to be better suited for control applications in robotics \cite{Sutton} (DDPG is used for instance in \cite{fei2020learn}). Because of its effectiveness in practice, observed by many authors, e.g. \cite{rl} for attitude control, we also compare with the on-policy 
Proximal Policy Optimisation \cite{PPO}, also used  for similar applications in 
\cite{simtoreal} and  
in \cite{stockholm}. 


DDPG is the historical method for continuous observation and action space applications to control, SAC and TD3 being improvements of DDPG. For instance, 
SAC regularizes the reward with an entropy term that is supposed to reduce the need to fine hyper-parameter tuning. 

Let us now describe the training mechanism: we call {\it query signal} the function describing the prescribed angular rates at any given time. 
We model this signal by a constant plateau, of magnitude chosen randomly between -0.6 and 0.6 radians per second, and duration chosen randomly between 0.1 and 1 second. We are training over a time window of 1 second (a training episode) during which the query signal is a constant plateau followed by a value of 0 until the end of the episode.
We chose to report on training where these query signals are used independently on pitch, roll and yaw. We tested joint queries as well but  do not report specifically the corresponding results since we observed no significant difference. 

Controls are updated every 0.03 seconds, and we simulate the full state of the quadrotor, using a Runge Kutta of order 4 on \Cref{eq:dynamic} with a time step of 0.01 seconds. 


The evaluation of the controller is made on similar query signals, but on time windows that last 20 seconds, with a query signal generated according to a more general class of queries (see below). 
Query signals on such longer time windows could also be considered for training : \cite{rl} refers to this approach as "continuous mode" and reports much poorer performance compared to the "episodic mode" with 1 second queries. We therefore decided to report only on episodic mode training. 


\begin{table}[htbp]
    \centering
    \begin{tabular}{@{}llrr@{}}
        \toprule
        Variable & Unit & Lower Bound & Higher Bound \\
        \midrule
        $z$ & \si{m} & -1000 & +inf\\
        $u$ & \si{m.s^{-1}} & {-30}  & {30}\\
        $v$ & \si{m.s^{-1}} &  {-30 }& {30}\\
        $w$ & \si{m.s^{-1}} & {-30}  & {30}\\ 
        $\phi$   & \si{\radian} & ${-\pi}$ & ${\pi}$\\
        $\theta$ & \si{\radian} & ${-\pi}$ & ${\pi}$\\
        $\psi$   & \si{\radian} & ${-\pi}$ & ${\pi}$\\
        $p$ & \si{\radian.s^{-1}} & ${-5\pi}$ & ${5\pi}$\\
        $q$ & \si{\radian.s^{-1}} & ${-5\pi}$ & ${5\pi}$\\
        $r$ & \si{\radian.s^{-1}} & ${-5\pi}$ & ${5\pi}$\\
        $cmd_\phi$   & PWM & -400 & 400\\
        $cmd_\theta$ & PWM & -400 & 400\\
        $cmd_\psi$   & PWM & -1000 & 1000\\
        $F$ & \si{\newton} & 0 & 52428\\
        \bottomrule
    \end{tabular}
    \caption{State and action space bounds}
    \label{params1}
\end{table}

Such query classes are characterised by three distributions $A$, $D$ and $S$ for respectively the amplitude and duration of stable plateaus, and the step amplitude between successive stable plateaus. These distributions are the same for each axis.
We define three different classes of queries (where U(a,b) denotes the Uniform distribution of support [a,b]):
    easy (A = U(-0.2, 0.2), D = U(0.5, 0.8), S = U(0, 0.3)), 
    medium (A = U(-0.4, 0.4), D = U(0.2, 0.5), S = U(0, 0.6)) and 
    hard (A = U(-0.6, 0.6), D = U(0.1, 0.2), S = U(0, 0.9)). 
Our query generator actually changes the joint distribution of amplitude and duration of stable plateaus by filtering out those queries which would make the roll, pitch and yaw go through singular values in the Euler angles description of the dynamics. 

The initial states are sampled in rather large intervals of values. These values as well as 
the maximal magnitudes of states are given in \Cref{params1}:






\section{Robust Signal Temporal Logic with Aggregates}
\label{sec:fastlr}

To formalize the behavioral properties of the closed-loop system we defined our own flavor of Signal Temporal Logic \cite{STL}. Our logic is mainly inspired by two preexisting works \cite{BeyondSTL} and \cite{TimeRobustness}. From \cite{BeyondSTL} we reuse the notion of aggregate operators over sliding windows and extend it with a robust quantitative semantics, where the original paper only defined a classic Boolean semantics for the language. From \cite{TimeRobustness} we reuse the idea of time-averaged robustness, and propose a new Until operator which combines both spatial robustness (instantaneous falsification margin at time t) and temporal robustness (robustness of the property to time delays over signals).

In this logic formulas are interpreted over piecewise-constant signals, whereas they were interpreted over piece\-wise-linear signal traces in \cite{BeyondSTL}. Our logic's semantics can however be extended to piecewise-linear signals without significant issue.

\subsection{Abstract Syntax}
\label{sec:fastlr-syntax}

Terms, Formulas and Aggregates:
\begin{align}
 \label{eq:terms1} \term \syndef& ~~~ c \synor x  \synor f(\term_1, \dots, \term_n) \\
 \label{eq:terms-ite}    &\synor \ite(\form, \term_1, \term_2) \\
 \label{eq:terms2}       &\synor \aggon{a,b} \aggr \\
 \label{eq:terms3}       &\synor \aggr \agguntil{a,b}{d_{\R}} \form \\
 \label{eq:terms4}       &\synor \term \tipuntil{a,b}{d_{\R}} \form \\
 \label{eq:terms5} \form \syndef& ~~~ \top \synor \bot \\
 \label{eq:terms6}       & \synor \term > 0 \\
 \label{eq:terms7}       & \synor \lnot \form \synor \form_1 \land \form_2 \synor \form_1 \lor \form_2 \\
 \label{eq:terms8}       &\synor \aggon{a,b} \aggb \\
 \label{eq:terms9}       &\synor \aggb \agguntil{a,b}{d_{\B}} \form \\
 \label{eq:terms10}      &\synor \form_1 \tipuntil{a,b}{d_{\B}} \form_2\\
 \label{eq:terms11}      &\synor \form_1 \avguntil{a,b} \form_2\\
 \label{eq:terms12} \aggr \syndef& ~ \aggmin~\term \synor \aggmax~\term \\
 \label{eq:terms13} \aggb  \syndef& ~ \aggforall~\form \synor \aggexists~\form
\end{align}
with $(a, b) \in \R^2$ and $a \leq b$, ${d_{\R}} \in \R$, ${d_{\B}} \in \B$.

A term  $\term$ is either: a constant $c$, a signal $x$ or a combinatorial
function $f$ applied to a number of terms (\ref{eq:terms1}); an if-then-else selection of a term based on a Boolean condition \ref{eq:terms-ite};
a value computed from a time interval $[a,b]$ using some numeric aggregate $\aggr$ (\ref{eq:terms2});
an ``aggregate until'' term $\aggr \agguntil{a,b}{d_{\R}} \form$ which computes a real value over
a time interval $[a,b]$ using a numeric aggregate $\aggr$ (\ref{eq:terms3}); or a ``time-point
until'' $\term \tipuntil{a,b}{d_{\R}} \form$, which samples the value of a term when
a formula becomes true (\ref{eq:terms4}).

A formula $\form$ is either: a logical constant true $\top$ or false $\bot$ (\ref{eq:terms5});
the comparison of a term to zero (\ref{eq:terms6}); 
the negation of a formula, or the conjunction or disjunction of a formula (\ref{eq:terms7}); 
an \emph{aggregate} computed from a time interval $[a,b]$ using some logic aggregate $\aggb$ (\ref{eq:terms8});
an \emph{aggregate until} formula $\aggb \agguntil{a,b}{d_{\B}} \form$ which computes a truth value
over a time interval $[a,b]$ using a logical aggregate $\aggb$ (\ref{eq:terms9});
a \emph{sample until} $\form \tipuntil{a,b}{d} \form$, which samples the value a formula when
some formula becomes true (\ref{eq:terms10});
or an \emph{average until} of a formula $\form_1$ computed over time interval $[a,b]$ until $\form_2$ becomes satisfied (\ref{eq:terms11}).
A \emph{numeric aggregate} $\aggr$ is either the min or max of a term $\term$ (\ref{eq:terms12}). 
A \emph{logic aggregate} $\aggb$ is either the Forall or Exists of a formula $\form$ (\ref{eq:terms13}). 

In addition to these core operators, the logic provides a number of derived operators defined in terms of the core operators.

The \emph{term lookup} operator is defined as follows:
\begin{equation}
  \lookup{a}{d} \term = \term \tipuntil{a,a}{d} \top
\end{equation}

The \emph{formula lookup} operator is defined as follows:
\begin{equation}
  \lookup{a}{d} \form = \form \tipuntil{a,a}{d} \top
\end{equation}

The original STL's Globally, Finally and Until operators are defined as
follows:

\begin{align}
\label{eq:fastlr-stl}
\finally{a,b} \form &= \aggon{a,b} ~\aggexists~ \form\\
\globally{a,b} \form &= \aggon{a,b} ~\aggforall~ \form\\
\form_1 \stluntil{a,b} \form_2 &= (\aggforall ~ \form_1) \agguntil{a,b}{\bot} \form_2
\end{align}

\subsection{Interpretation Structures}
\label{sec:fastlr-interpretation-structures}

Terms and formulas are interpreted over total piecewise-constant functions $\trace: \R \rightarrow \R^n$, which assign a value to a tuple of signals $X = (x_1,\dots,x_n)$ of size $n$ at any time $t \in \R$.
However, for practical reasons we only consider total piecewise-constant functions defined by a finite sequence of \emph{breakpoints}: 
\[
\mathit{Bkpts} = \llbracket (t_j, X_j) ~|~ j\in [0, M-1], (t_j, X_j) \in (\R,\R^n),  \rrbracket
\]
where $t_j < t_{j+1}$ for all $j$, and by a default value $X_d \in \R^n$, as follows:

\begin{equation}
\trace(t) = 
\begin{cases}
X_d & \text{ if } t \in (-\infty, 0) \\
X_j & \text{ if } t \in [t_j, t_{j+1}) \\
X_{M-1} & \text{ if } t \in [t_{M-1}, +\infty) \\
\end{cases}
\end{equation}

We use the following notations:

\begin{itemize}
\item $\timesteps_{\trace} = \llbracket t_j ~|~ j \in [0,M-1] \rrbracket$ is its sequence of timesteps,
\item $\trace(x_i, t)$, by abuse of notation, is the $i^{th}$ coordinate of $\trace(t)$, i.e. the value of signal $x_i$ at time $t$.
\end{itemize}

\subsection{Standard Semantics}\label{sec:semantics}

\subsubsection{Term Semantics}\label{sec:term-semantics}

Assuming some fixed trace $\trace$, the interpretation function for terms

\begin{equation}
  \sem{}: \term \rightarrow \R \rightarrow \R
\end{equation}

is defined inductively as follows:

\begin{align}
  \sem{c}(t) &= c \\
  \sem{x_i}(t) &= \trace(x_i,t) \\
  \sem{f(\term_1, \dots, \term_n)}(t) &= f(\sem{\term_1}(t), \dots, \sem{\term_n}(t)) \\
  \sem{\ite(\form, \term_1, \term_2)}(t) &= 
    \begin{cases}
        \sem{\term_1}(t) & \text{if} ~ \sem{\form}(t) \\
        \sem{\term_2}(t) & \text{otherwise} \\
    \end{cases}\\
  \sem{\aggon{a,b} \aggr}(t) &= \sem{\aggr}([t+a,t+b]) \\
  \sem{\aggr \agguntil{a,b}{d} \form}(t) &=
                                           \begin{cases}
                                             \sem{\aggr}([t,t']), \text{where} ~ \\
                                             ~ ~ t' \in [t+a,t+b]
                                             ~ \text{smallest} \\ ~~~\text{instant st.} ~ \sem{\form}(t') = \top \\
                                             d~\text{if no such $t'$ exists}.
                                           \end{cases}\\
  \sem{\term \tipuntil{a,b}{d} \form}(t) &=
                                           \begin{cases}
                                             \sem{\term}(t'), \text{where} ~ \\
                                             ~ ~ t' \in [t+a,t+b] 
                                             ~ \text{smallest} \\ ~~~\text{instant st.} ~ \sem{\form}(t') = \top \\
                                             d~\text{if no such $t'$ exists}.
                                           \end{cases}
\end{align}

The semantics of numeric aggregates is defined over intervals as follows:
\begin{align}
  \sem{\aggmin ~ \tau}([a, b]) &= \mathrm{min}_{t \in [a,b] \cap \timesteps_{\trace}}(\sem{\form}(t)) \\
  \sem{\aggmax ~ \tau}([a, b]) &= \mathrm{max}_{t \in [a,b] \cap \timesteps_{\trace}}(\sem{\form}(t))
\end{align}

Since traces are total piecewise-constant functions, defined by a finite number of samples, and all operators have default values and are hence total, interpretation functions are also total function, and evaluating a numeric or logic aggregates requires inspecting only a finite number of timesteps and yields an exact result.


\subsubsection{Formula Semantics}\label{sec:formula-semantics}

Assuming a fixed trace $\trace$, formula semantics is given by the function:
\begin{equation}
  \sem{}: \form \rightarrow \R \rightarrow \B
\end{equation}

defined inductively as follows:

\begin{align}
  \sem{\top}(t) &= \top \\
  \sem{\bot}(t) &= \bot \\
  \sem{\term > 0}(t) &= \sem{\term}(t) > 0 \\
  \sem{\lnot \form}(t) &= \top ~\text{iff}~ \sem{\form}(t) = \bot \\
  \sem{\form_1 \land \form_2}(t) &= \top ~\left\{\begin{array}{l} 
  \text{iff}~ \sem{\form_1}(t) = \top \\
  ~\text{and}~  \sem{\form_2}(t) = \top\\
  \end{array}\right. \\
  \sem{\form_1 \lor \form_2}(t) &= \top ~\left\{\begin{array}{l}
  \text{iff}~ \sem{\form_1}(t) = \top \\
  ~\text{or}~  \sem{\form_2}(t) = \top\end{array}\right.\\
  \sem{\aggon{a,b} \aggb}(t) &= \sem{\aggb}([t+a,t+b]) \\
  \sem{\aggb \agguntil{a,b}{b} \form}(t) &=
                                           \begin{cases}
                                             \sem{\aggb}([t,t']), ~\text{where}~ t' \\ \ \ \text{in }  [t+a,t+b]
                                             \text{ smallest } \\ \ \ \text{timestep st. } ~ \sem{\form}(t') = \top \\
                                             b~\text{if no such $t'$ exists.}
                                           \end{cases}\\
  \sem{\form_1 \tipuntil{a,b}{b} \form_2}(t) &= \begin{cases}
                                             \sem{\form_1}([t,t']), ~\text{where}~ t' \\ \ \ \text{in }  [t+a,t+b]
                                             \text{ smallest } \\ \ \ \text{timestep st. } ~ \sem{\form_2}(t') = \top \\
                                             b~\text{if no such $t'$ exists.}
                                           \end{cases}
\end{align}

The classic semantics for the \emph{average until} operator $\form_1 \avguntil{a,b} \form_2$ is defined exactly as the original \emph{STL Until} semantics.

The semantics of logic aggregates is defined over intervals $[a,b]$ as follows:
\begin{align}\label{eq:sem-logagg}
  \sem{\aggforall ~ \phi}([a,b]) &= \bigwedge_{t \in [a,b]}{\sem{\phi}(t)}\\
  \sem{\aggexists ~ \phi}([a,b]) &= \bigvee_{t \in [a,b]}{\sem{\phi}(t)}
\end{align}

\subsection{Robust semantics}\label{sec:robust-semantics}

The robust semantics
\begin{equation}
  \robust{}: \form \rightarrow \R \rightarrow \R
\end{equation}
only concerns Boolean formulas, and is defined inductively as follows:

\begin{align}\label{eq:robust-form}
  \robust{\top}(t) &= +\infty \\
  \robust{\bot}(t) &= -\infty \\
  \robust{\term > 0}(t) &= \sem{\term}(t)\\ 
  \robust{\lnot \form}(t) &= -\robust{\form}(t)\\
  \robust{\form_1 \land \form_2}(t) &= \mathrm{min}(\robust{\form_1}(t), \robust{\form_2}(t))\\
  \robust{\form_1 \lor \form_2}(t) &= \mathrm{max}(\robust{\form_1}(t), \robust{\form_2}(t))\\
  \robust{\aggon{a,b} \aggb}(t) &= \robust{\aggb}([t+a,t+b]) \\
  \robust{\aggb \agguntil{a,b}{b} \form}(t) &=
                                              \begin{cases}
                                         \robust{\aggb}([t,t'])~  \text{where}~ t' \text{ in }  \\ \ \ [t+a, t+b] 
                                    \text{ is the smallest } \\ \ \ \text{ timestep st. }  \sem{\form}(t') \\
                                                \robust{b}~\text{if no such $t'$ exits.}
                                           \end{cases}\\
  \robust{\form_1 \tipuntil{a,b}{b} \form_2}(t) &=
                                           \begin{cases}
                                              \robust{\form_1}(t') ~
                                                \text{where}~ t' \text{ in }  \\ \ \ [t+a, t+b] 
                                    \text{ is the smallest } \\ \ \ \text{ timestep st. }  \sem{\form_2}(t') \\
                                                \robust{b}~\text{if no such $t'$ exits.}
                                           \end{cases}\\
  \robust{\form_1 \avguntil{a,b} \form_2}(t) &=
                                           \begin{cases}
                                                (b-t')*\robust{\aggon{t,t'} \aggforall ~ \form_1}(0)\\
                                              \ \   \text{where}~ t' \in [t+a,t+b] ~ \\ \ \ \text{is the smallest timestep } \\
                                              \ \ \text{ st. } ~ \sem{\form_{2}}(t')\\
                                                -\infty ~ \text{if no such $t'$ exists.}
                                  \end{cases}\\  
\end{align}

The robust semantics for logic aggregates is defined as follows:
\begin{align}\label{eq:robust-logagg}
  \robust{\aggforall ~ \phi}([a,b]) &= \mathrm{min}_{t \in [a,b]}{\robust{\phi}(t)}\\
  \robust{\aggexists ~ \phi}([a,b]) &= \mathrm{max}_{t \in [a,b]}{\robust{\phi}(t)}
\end{align}

The robust interpretation of terms is just their standard interpretation,
except for \emph{timepoint until} and \emph{aggregate until} operators where instead of recursing on the standard interpretation of Boolean formulas, we recurse on their robust interpretation.

\subsection{Implementation}\label{sec:fastlr-implementation}

We implemented a code generator for the logic, which generates highly efficient python code allowing to compute the standard and robust semantics of STL formulas on traces.
Given an STL formula (or term) as input, the code generator produces a Python 3.x class definition which implements the standard and robust semantics evaluation rules for that formula. The class takes a trace as constructor argument (i.e. a piecewise constant function specified a sequence of breakpoints and a default value as defined in \cref{sec:fastlr-interpretation-structures}), and offers an \texttt{eval} method allowing to compute the standard or robust semantics of the formula at any time step. 

The generated code uses a number of techniques for efficiency: 
\begin{itemize}
\item Constant folding,

\item When translating a specification containing several formulas and terms, the code generator implements common subformulas and subterms sharing between all toplevel formulas.

\item We leverage the fact that in practice, a same formula will be evaluated on sequences of strictly increasing timesteps, and use an incremental method for the evaluation of sliding window aggregates: when evaluating a Max aggregate (resp. Min, Forall or Exists) at time $t$, the aggregate term is evaluated on interval $[t+a,t+b]$ and we cache the result $(t_{Max}, x_{Max})$, indicating at which instant the Max value was reached in [t+a,t+b]. When the aggregate is evaluated again at $t' > t$, we distinguish the following cases:
\begin{itemize}
\item if $t_r \in [t'+a, t+b]$, we evaluate the aggregate on $[t+b, t'+b]$ and return $Max(x_Max, x_Max')$, cache new result, 
\item if $t_r \in [t+a, t'+a]$, we evaluate the aggregate on the full window $[t'+a, t'+b]$ and cache the result.
\end{itemize}
\item Last, the generated code uses numpy arrays exclusively and contains Numba annotations for all data structures and classes, allowing to use Numba to JIT the evaluation code using LLVM. This JIT optimization provides a 10x to 20x performance boost over the interpreted python version.
\end{itemize}



\section{Formal performance criteria}
\label{sec:perf-observers}

Designing a controller for a specific application requires balancing multiple criteria such as rising time, overshoot, steady error, etc. 
In order to quantify rigorously the performance of the learned controller, we formalized requirements using the logic presented in \cref{sec:fastlr}. 

A first set of formulae allows to identify instants when a query signal $q$ becomes stable for $T$ time units, 
and whether $q$ goes up or down at any instant (with $\epsilon$ and $d$ two small constants), and the step size:
\begin{align}
       \stable(q) &= (\On_{[0, T]} ~ \MaxAgg ~ q) - (\On_{[0, T]} ~ \MinAgg ~ q) < d\\
\becomesstable(q) &= (D_{-\epsilon}^\bot ~ \lnot \stable(q)) \land \stable(q)\\
           \up(q) &= q - (D_{-\epsilon}^0 ~ q) > 0 \\
         \down(q) &= q - (D_{-\epsilon}^0 ~ q) \leq 0 \\
        \stepSize(q) &= \ite(\becomesstable(q), q - D_{-\epsilon}^0{q}, 0) 
\end{align}

We consider an angular rate signal $x$ as acceptable if it does not overshoot a stable query $q$
by more than $\alpha\%$ of the step size on $[0, T_1]$, and does not stray away from a stable query $q$ 
by more than $\beta\%$ of the step size on $[T_1,T]$:
\begin{multline}
\becomesstable(q) \land \up(q)   \implies \\ \On_{[0, T_1]} ~ \MaxAgg ~  (x - q)  < \alpha \stepSize(q) 
\label{eq:overshootup}
\end{multline}
\begin{multline}
\becomesstable(q) \land \down(q) \implies \\ \On_{[0, T_1]} ~
 \MaxAgg ~  (q - x)  < \alpha \stepSize(q) 
\label{eq:overshootdown}
\end{multline}
\begin{multline}
\becomesstable(q)                \implies \On_{[T_1, T]} ~ \MaxAgg ~ \|x - q\| < \beta \stepSize(q)
\label{eq:offset}
\end{multline}

We define the rising time $\mathit{RT}$ as the time it takes for $x$ to first reach $q$ within $\gamma\%$:
\begin{equation}
\ite(\becomesstable(q),  t - (t ~ U_{[0,T]}^{+\infty} ~ \|(x-q)\| < \gamma q), +\infty)
\label{eq:rising}
\end{equation}

\begin{figure}[htbp]
    \centering
    \includegraphics[width=\linewidth]{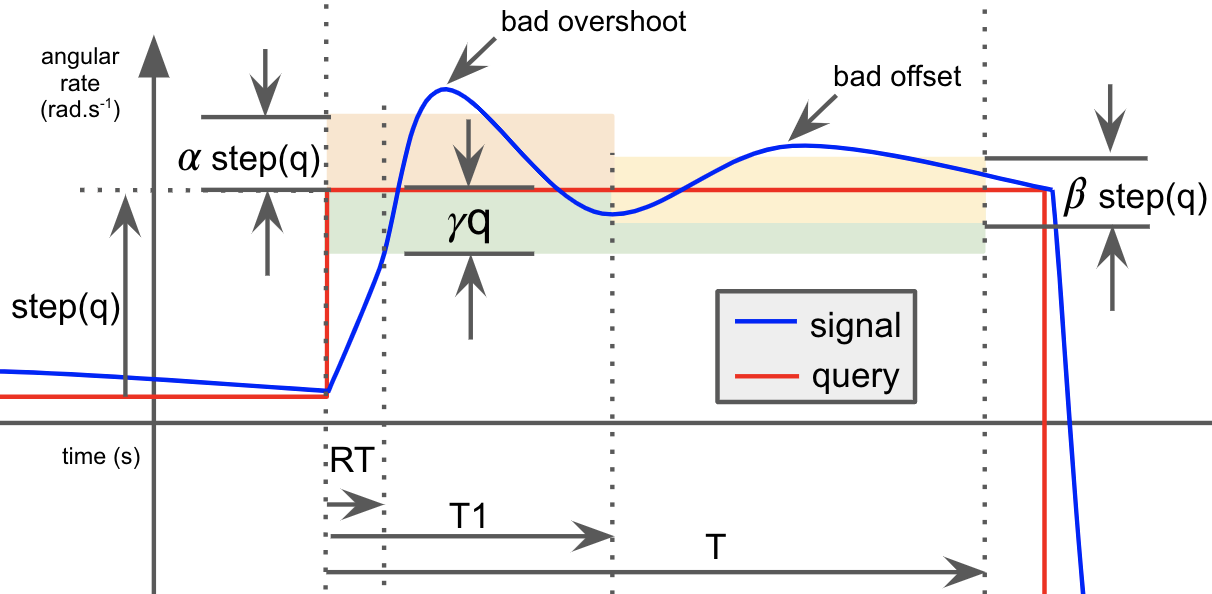}
    \caption{Property parameters $T_1$, $T$, $RT$, $\alpha$, $\beta$, $\gamma$.}
    \label{fig:observer-params}
\end{figure}

\Cref{fig:observer-params} illustrates the formalised notions and parameters.

Using observers code generated from these specifications, we compute statistics about property violations 
and associated robustness margins on angular rate signals and queries on pitch, yaw and roll axis of the system, 
acquired at regular intervals during the training of the controller. For evaluation each property $P(x,q)$ is wrapped 
in a \emph{globally} modality over the episode length yielding $G_{[0,\mathit{episode\_length}]} ~ P(x,q)$.
Automating the computation of these behavioral metrics is essential in allowing to scale up the hyper-parameter
space exploration and identify the best controller according to objective measurements.

\section{Experimental setup}

\label{sec:experiments}

\subsection{Implementation}

We have developed a platform\footnote{The full code is available as open source at \url{https://github.com/uber-research/rl-controller-verification}.} 
with the purpose of running experiments in a reproducible and scalable way, becoming an integration layer between the different moving parts in both training and testing.
From a technological standpoint the platform is based on the Stable Baselines 2.7.0 reinforcement learning library \cite{stable-baselines} itself based on Tensorflow \cite{abadi2016tensorflow}, all of our code is in Python and we used Bazel \cite{Bazel} as build system. We used Tensorboard to monitor losses and the internal dynamics of the neural networks during the training.

One intermediate goal was to explore the large combinatorial hyperparameter space efficiently, to be able to identify the best hyperparameters values with respect to the STL metrics we defined and to get a better understanding of their impact.

With 4 different algorithms, 20 possible configurations for the network architecture and 3 sets of observed states, our hyperparameters space contains a total of 240 points that need to be trained and tested. The corresponding jobs are dispatched on our Kubernetes cluster \cite{Kubernetes} where they can run in parallel. Disposing of 1 vCPU on the Cascade Lake platform (base frequency of 2.8 GHz), the 3 millions iterations of a single training job take between 3 and 8 hours to complete. The cluster autoscales with the workload and allowed us to run 1,200 hours worth of training in half a day.

The container images that end up running on the cluster are created, uploaded and finally dispatched in a reproducible manner thanks to the Bazel rules of our Research Platform. Those rules are built on top of the Bazel Image Container Rules \cite{bazel-rules-docker} and the Bazel Kubernetes Rules \cite{bazel-rules-k8s} and specially designed to generate all the experiment jobs of the hyperparameters analysis.

The training and testing results are automatically uploaded on our cloud storage where they can be browsed for quick inspections, or fed as input for the next pipeline stage.
We saved 30 checkpoints per experiment (each file containing 100k training iterations weights between 10KB and 100KB). Including the TensorFlow logs, the training results amount to over 100GB of data.

Each of the 30 $\times$ 240 checkpoints was then evaluated on 100 queries computed by the Query Generator, producing the same number of concrete traces representing the commands and the states over the whole episode. Each set of such traces is about 600k hence it yields total of 60MB per checkpoint. 
Finally each of the 30 $\times$ 240 $\times$ 100 traces was evaluated with STL properties observer to compute synthetic metrics: aggregating the 100 traces of a single checkpoint produced a 150KB file and required approximately 45 minutes.
The checkpoint-specific CSV files were further aggregated in experiment-specific and round-specific checkpoints for  final visual inspection.

\subsection{Interactive browsing of the experiments database}

\begin{figure*}
\includegraphics[width=\linewidth]{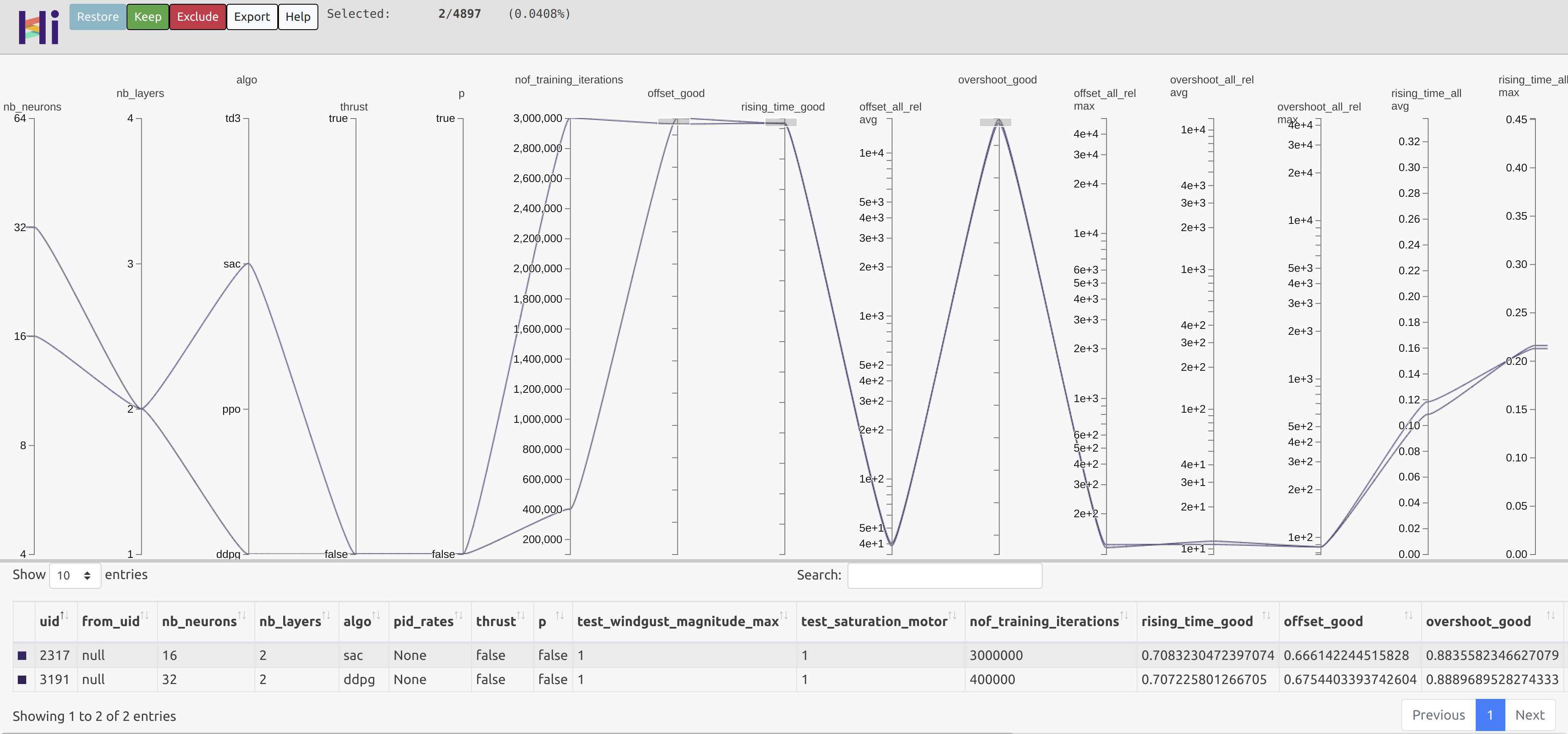}
\caption{Hiplot interactive session}
\label{fig:hiplot}
\end{figure*}

We want to understand what correlations exist between controller performance and the way it has been trained, and for this, we used Hiplot
\cite{hiplot} for browsing through the enormous number of parameters and data generated. 
We show in \Cref{fig:hiplot} how we used Hiplot in an interactive manner for verifying our hypotheses. Each parameter and performance measure is represented by a column in the graph generated by Hiplot from our database. For each parameter, either fixed or free, choosing intervals of values for each performance measure creates lines that link parameter values to performance values within the chosen intervals. The number of entries in the database (i.e. the number of controllers) that satisfy the constraints is also shown, as well as the table of all their corresponding parameters and performance values. 

For instance, we used Hiplot to select the "best" networks,  filtering the data set of controllers, only retaining the ones with better success in offset, overshoot and rising times altogether, with respect to the best PIDs.
This resulted in two neural nets with much better performances than the PIDs on offset and on rising time, as we will discuss in \Cref{sec:exp-results}.

\section{Experimental results}
\label{sec:exp-results}
\subsection{Performance metrics}

Each controller is evaluated on a hundred evaluation episodes using STL observers defined in \Cref{eq:overshootup,eq:overshootdown,eq:offset,eq:rising},
where parameters are set to $\alpha=10\%$, $\beta=5\%$ and $\gamma=5\%$, $T=0.5s$, $T_1=0.25s$, $\epsilon = 0.01s$, $d=0.005$.
For each evaluation episode the following statistics are computed over all stable query plateaus: 
\begin{itemize}
\item average and maximum overshoot percentage relative to the query step size, 
\item average and maximum offset percentage relative to the query step size, 
\item average and maximum rising time values in seconds (only for plateaus where the signal actually reaches $\gamma\%$ of the query within $[0,T]$).
\end{itemize}

For each metric (overshoot, offset, rising time), we compute the \emph{success percentage} {\% OK}, i.e. the percentage of stable plateaus of the episode for which the controller behaviour satisfies the specification.

Then, episode-level statistics are further averaged, yielding results presented in the tables of the following sections, where columns represent:
\begin{itemize}
    \item avg (resp. max) overshoot: is the per-episode-average of the average (resp. maximum) overshoot values,
    \item avg (resp. max) offset: is the per-episode-average of the average (resp. maximum) offset values,
    \item avg (resp. max) rising time: is the per-episode-average of the average (resp. max) rising time, 
    \item \% OK offset (resp. overshoot, rising time): is the per-episode-average of the success percentage for the offset (resp overshoot, rising time) metric. 
\end{itemize}

\subsection{Performance of nominal-trained networks in nominal test case}
\label{sec:expnominal}




\subsubsection{Overall best performance comparison}
\begin{figure}
\begin{center}
\epsfig{file=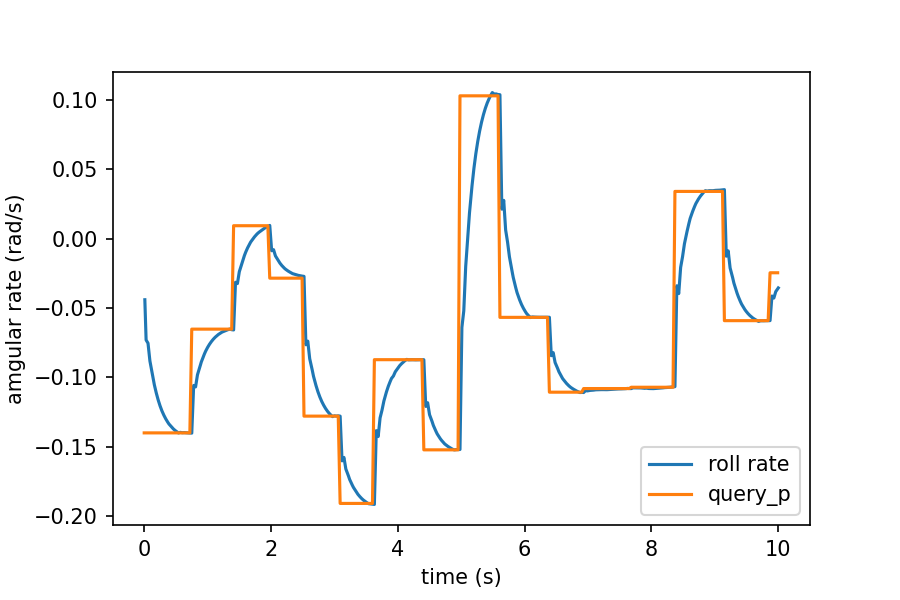,clip=,width=0.45\textwidth}
\caption{PID2 controller query tracking}
\end{center}
\label{fig:pid_traces}
\end{figure}

The PID performance metrics in the nominal case are reported in the first two lines of \Cref{fig:pidnnnominal} to serve as a reference point for neural controller evaluation.
Examples of query tracking behavior are given in \Cref{fig:pid_traces} for reference.

PID2 reaches within 5\% of the target state for about 70\% of the queries, and is relatively slow with an average rising time of 0.44s. PID1 in comparison reaches within 5\% of the target state for only 8\% of the queries, with a (very slightly) better rising time. Overshoot success rates are really good for both PIDs (95-100\% OK). Offset success rates are bad (1-3\% OK), due to their slow convergence.
We will hence use PID2 as a reference for discussing neural controller performance.

\begin{figure}
\begin{center}
\includegraphics[width=0.9\linewidth]{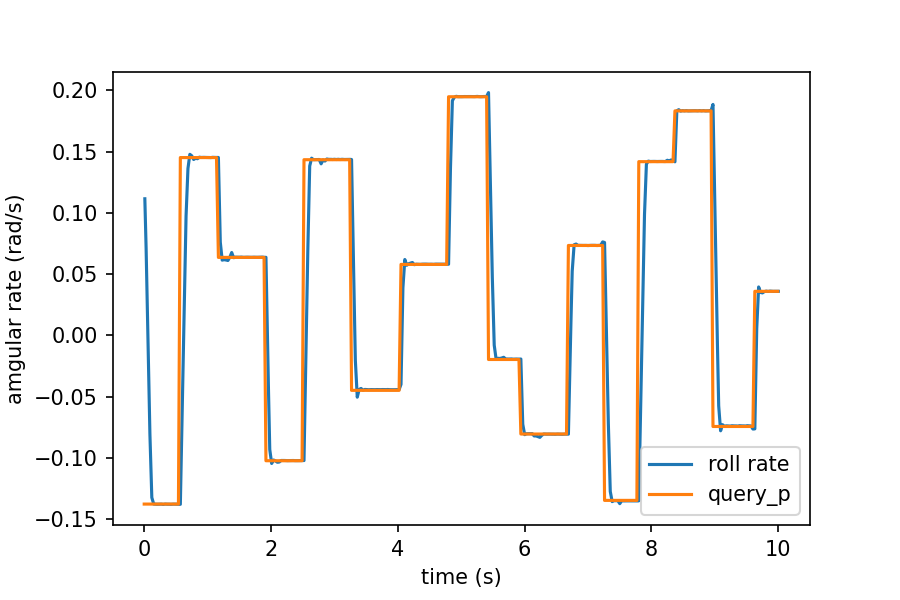}
\caption{Neural controller behaviour (sac, 2 layers, 16 neurons per layer, 3M iterations)}
\label{fig:samplespike}
\end{center}
\end{figure}

The comparison between the best networks and the PIDs is also reported in \Cref{fig:pidnnnominal}. 
\begin{csvtable*}{all_algos_pid_new.csv}[table head=\toprule algo & OK & OK & OK & avg & avg & avg & max & max & max \\ & rising t. & off. & overshoot & rising t.& off. &  overshoot & rising t. & off. & overshoot \\\midrule, before reading=\footnotesize]
\caption{PIDs and overall best networks performance (all in \% except rising t. in seconds)}
\label{fig:pidnnnominal}
\end{csvtable*}
We see that our neural nets provide much quicker controls, with an average rising time of about a fourth to a fifth of the rising time for the two PIDs, although with a negligible offset. 
This is at the expense of a slightly less good performance on the maximum overshoot at least for SAC and DDPG trained networks, with respect to PID2 (our neural nets are still much better than PID1). Results are far less good, in particular concerning overshoots, with PPO and TD3 trained networks. 
This is also visible when comparing signals between  \Cref{fig:samplespike} and \Cref{fig:pid_traces}. Somehow, neural nets exhibit extreme reactivity as well as good asymptotic convergence, but show some very short-lived "spikes", as in the sample trajectory shown in \Cref{fig:samplespike}. 

When we filter the neural nets meeting or exceeding the performances of PID2, many networks remain, among which  
the best are:
\begin{itemize}
\item DDPG $64\times 64\times 64\times 64$ trained for 1,500,000  iterations (and also DDPG $32\times 32$, 400,000 iterations) 
on the three-dimensional observation space $(p-p_{sp}, q-q_{sp},r-r_{sp})$
%
    \item SAC $32\times 32 \times 32 \times 32$ (and SAC $32 \times 32$ and $16\times 16$ trained for 3,000,000 iterations coming very close) trained for  2,900,000 iterations on the same three-dimensional observation space
\end{itemize}

\subsubsection{Training algorithm influence}

We observe in \Cref{fig:pidnnnominal}
that PPO and TD3 do not show as good performance as SAC (and even DDPG), moderating the conclusion of \cite{rl}, and the common belief that TD3 should improve performance of neural net control.
We have for now no explanation for this, largely because we have not been able (which is also the case in  \cite{rl}) to get rid of the overshoot spikes, even using SAC which does some amount of regularization, or TD3 which should lead to more stable solutions, potentially at the expense of a slower convergence rate. In terms of optimal control, if the neural net controller were trained with correctness objectives\footnote{Future work to cope with this phenomenon includes improving the reward function using our STL observers, and adding some more regularization during training.}, these spikes would certainly be much smaller and appear only at the very beginning of plateaus. 

\subsubsection{Convergence of the training algorithms}

\begin{figure}
\begin{center}
\includegraphics[width=0.9\linewidth]{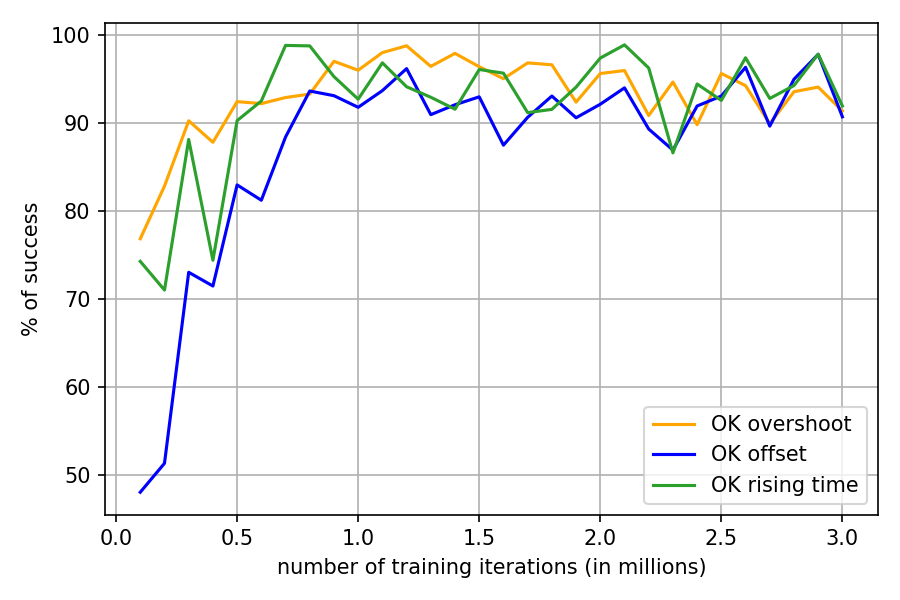}
\caption{Performance of SAC 32x32 on dim 3 observation space trained neural nets w.r.t. the number of iterations}
\label{table:perfsaciterations}
\end{center}
\end{figure}

 We show in \Cref{table:perfsaciterations} the evolution of the three main performance measures, the OK overshoot, OK offset and OK rising time, for one of the best network architecture and training algorithm, SAC $32 \times 32$ neurons. 
 The three metrics improve quickly and almost stabilize in the first 1,000,000 iterations. 
 


\subsubsection{Observation state influence}


Of course, for a given number of iterations, smaller-dimensional  observation states yield  better quality of the sampling. 
Still, we observe that using a Markovian state or the simpler three-dimensional state space $(err_p,err_q,err_r)$ does not change significantly the performance of the best neural nets obtained, see \Cref{fig:influencedim}, although the 3-dimensio\-nal observation space gives slightly better performance overall. In fact, we even get a worse performance with the 7-dimensional full state, mostly because of the difficulty to sample this higher dimensional space, and identify the subtle second-order effects of some of these states on angular rates.  

\begin{csvtable*}{influence_nof_states3_new.csv}[table head=\toprule algo & dim & OK & OK & OK & avg & avg & avg & max & max & max \\ &  & rising t. & off. & overshoot & rising t. & off. & overshoot & rising t. & off. & overshoot \\\midrule, before reading=\footnotesize]
\caption{Influence of the observable space dimension (all in \% except rising t. in seconds)}
\label{fig:influencedim}
\end{csvtable*}

\subsubsection{Neural net architecture influence}

\begin{figure}
\centering
\includegraphics[width=0.9\linewidth]{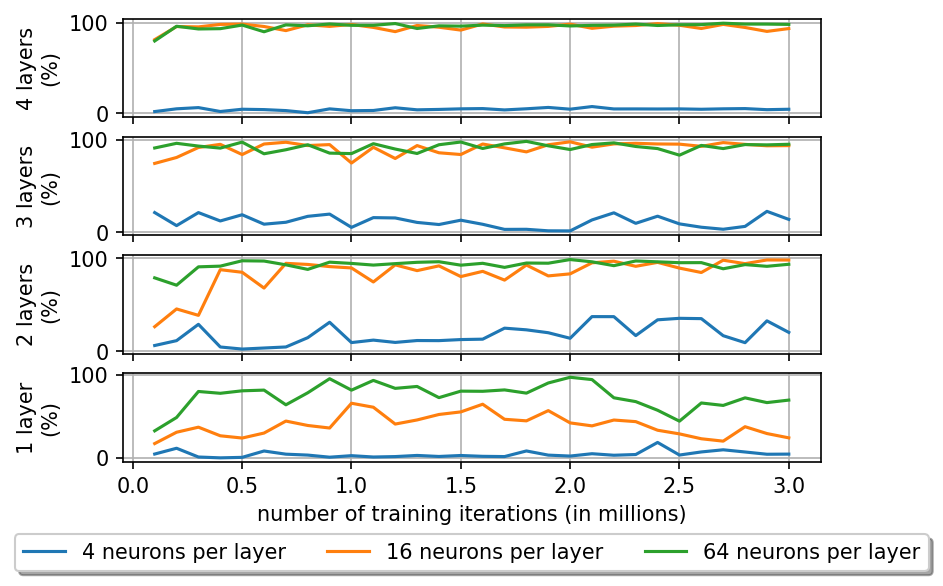}
\caption{OK rising t. for our best SAC network wrt number of training iterations for different architectures}
\label{fig:architectures-iterations}
\end{figure}

First, we observe that almost none of the single-layer neural nets seem to converge to a correct controller (see e.g. \Cref{fig:architectures-iterations}). At 64 neurons, 1 hidden layer networks seem to exhibit some good behaviour, but still far from any of the e.g. two-layers neural nets. 

Still, 3-layers and even 4-layers networks do not seem to exhibit much better behaviour than the "best" 2-layers networks, with 16 or 32 neurons each, although they converge faster. 
Recently Sinha et al. in~\cite{sinha2020d2rl} empirically observed the performance of SAC have a peak using 2 layers MLP and their explanation for this result relies on the Data Processing Inequality hence the fact that mutual information between layers decreases with depth. This will have to be further investigated in our framework. 

\subsection{Performance of nominal-trained networks in non-nominal test cases}

\label{sec:expnonnominal}

We now assess the robustness of our PIDs and "best" neural nets (trained in nominal situations as discussed in  \Cref{sec:expnominal}) to perturbed, non-nominal conditions, without training the neural nets nor changing the gains of PIDs to cope specifically for the new situation. 
We report the same performance measures as the ones used in the nominal case, in the test cases where a perturbation can happen, at the start of any new plateau along the 20 second episodes that we are observing (which can contain about 30 different target angular states, or plateaus, to reach within a short time). We take maxima and averages of these measures on 100 such queries as before. 

\subsubsection{Robustness to partial motor failures}

We report in \Cref{table:robustmotorfail} 
results where the perturbation is  a partial power loss of motor 1, down to 80\% of its maximal power.

For this case of partial motor failure, our best SAC trained neural net behaves much better than our two PIDs: it keeps on reaching plateaus within 0.5 seconds for about 94\% of the time, whereas even the best PID goes down to less than 60\% success rate. Our network is even better when it comes to satisfying offset constraints (82\% of the time) whereas the PIDs almost never comply. Performances concerning overshoot are comparable, even though the PIDs are very slightly better, but this only concerns cases where PIDs actually reach the target state, which is the case much less often. Essentially, the best neural nets that have been trained under nominal conditions show very little degradation of performance when a partial failure occurs.  





\subsubsection{Robustness to wind gusts}

We present in \Cref{table:robustwindgust} results where the perturbation is the occurence of randomly chosen wind gusts (as described in \Cref{sec:aero}) of magnitude up to 10 $m.s^{-1}$ from any fixed direction in the inertial frame.

\begin{csvtable*}{non_nominal_test_new.csv}[table head=\toprule mode & algo & OK & OK & OK & avg & avg & avg & max & max & max \\ &  & rising t. & off. & overshoot & rising t. & off. & overshoot & rising t. & off. & overshoot \\\midrule, before reading=\footnotesize\hspace*{-5mm}]
\caption{Robustness of the best networks and PIDs in case of wind gusts and motor saturation (all in \% except rising t. in seconds)}
\label{table:robustnonnominal}
\label{table:robustwindgust}
\label{table:robustmotorfail}
\end{csvtable*}

The PIDs and the neural nets exhibit the same kind of minor loss of performance, and the nominal trained neural nets are still far superior to the two PIDs. 

\subsection{Performance of non-nominal-trained networks}
\label{sec:perfnonnominaltrained}

\begin{csvtable*}{training_s_test_snew.csv}[table head=\toprule algo & OK & OK & OK & avg & avg & avg & max & max & max \\ & rising t. & off. & overshoot & rising t. & off. & overshoot & rising t. & off. & overshoot \\\midrule, before reading=\footnotesize]
\caption{Best networks trained for partial motor failures, tested under potential motor failures situations (all in \% except rising t. in seconds)}
\label{table:failnonnomnonnom}
\end{csvtable*}

\subsubsection{Training under partial motor failures}
In what follows, we train the attitude controller to sustain 
partial motor failures
adding the magnitude of the power loss (1 extra dimension) to the observation states discussed in \Cref{sec:Markov}. We report the performance measures obtained in the non-nominal case in \Cref{table:failnonnomnonnom}. 
The concern one may have is that, training the neural net in more various conditions (nominal and non-nominal), the resulting controller may exhibit lower performance. We thus report the same performance measures for neural nets trained with potential motor failures, in nominal situations, e.g. when no power loss happens, see  \Cref{table:failnonnomnom}

We see that we still achieve much better performance than PIDs, but that we are only similar and even slightly worse than the neural nets trained in nominal conditions, both in nominal conditions (compare \Cref{table:failnonnomnom} to  \Cref{fig:pidnnnominal})  and in non-nominal conditions (compare  \Cref{table:failnonnomnonnom} to \Cref{table:robustmotorfail}). Understanding this non intuitive behaviour and improving the training in this case is left for future work.

\begin{csvtable*}{training_s_test_nnew.csv}[table head=\toprule algo & OK & OK & OK & avg & avg & avg & max & max & max \\ & rising t. & off. & overshoot & rising t. & off. & overshoot & rising t. & off. & overshoot \\\midrule, before reading=\small]
\caption{Performance of best networks trained with potential motor failures, and tested in nominal situations (all in \% except rising t. in seconds)}
\label{table:failnonnomnom}
\end{csvtable*}

\subsubsection{Training under wind gusts}

In what follows, we train the attitude controller to sustain wind gusts up to 10 m.s$^{-1}$ in any direction, 
adding to the observation states we discussed in \Cref{sec:Markov} the wind gust magnitude and directions (4 additional dimensions) plus the linear velocities of the quadcopter ($u$, $v$ and $w$, 3 additional dimensions) since they are necessary for determining the relative wind velocity. 

We report the performance measures that we get in the non-nominal case in \Cref{table:windnonnomnonnom} and in the nominal case in \Cref{table:windnonnomnom}.
\begin{csvtable*}{training_w_test_wnew.csv}[table head=\toprule algo & OK & OK & OK & avg & avg & avg & max & max & max \\ & rising t. & off. & overshoot & rising t. & off. & overshoot & rising t. & off. & overshoot \\\midrule, before reading=\small]
\caption{Best networks trained for wind gusts conditions, tested under wind gusts conditions (all in \% except rising t. in seconds)}
\label{table:windnonnomnonnom}
\end{csvtable*}

We see that the SAC and DDPG controller trained with potential wind gusts still behave about as well as the nominal controller (compare \Cref{table:windnonnomnom} to \Cref{fig:pidnnnominal}). Surprisingly, the best (SAC) network behaves slightly worse than the nominal-trained SAC network under wind gusts (compare \Cref{table:windnonnomnonnom} to \Cref{table:robustwindgust}), where we can see a slight drop of performance in e.g. {\em OK off.} and {\em OK overshoot}: it does not seem to be able to learn correctly how to stay close enough to the target plateau, in some cases. 

\begin{csvtable*}[h!]{training_w_test_nnew.csv}[table head=\toprule algo & OK & OK & OK & avg & avg & avg & max & max & max \\ & rising t. & off. & overshoot & rising t. & off. & overshoot & rising t. & off. & overshoot \\\midrule, before reading=\small]
\caption{Best networks trained for wind gusts conditions, tested in nominal conditions (all in \% except rising t. in seconds)}
\label{table:windnonnomnom}
\end{csvtable*}




\section{Lessons learned}

\label{sec:lessonslearned}



\paragraph{Sampling}

First, we observed that we should restrict to a “good” subspace of the (full quadcopter) states that is sufficiently low dimensional for efficient sampling and such that it avoids potentially spurious correlations, while still providing sufficient information for learning. For instance, in the nominal case, the observation space $(err_p,err_q,err_r)$ was found to be the optimal choice. 
Training depends of course on sampling data, that has to be done on representative data, and on sampling initial states in a large enough space. In order to do this, for better results, we developed a specific query generator, and we sampled initial states in quite large spaces. 

\paragraph{Training algorithms}

SAC gives very good results as expected. It is most probably more efficient due to entropy regularization that partially cancels spurious correlations, but this  has still to be confirmed in more general situations. 
A lesson for us was that TD3 was not behaving as well as expected. Our current guess is that TD3 suffers from too much bias on the Q-function estimation at some point in our training environment, or that TD3 needs many more iterations to converge in our case due to bad exploration performance. Recent papers have suggested that action clipping in TD3 can result in poor exploration performance on problems with bounded action spaces (actions on the boundary are too frequently sampled) which has been shown to be remedied by the entropy regularization of SAC or other output scaling and replay buffer sampling approaches that simulate entropy regularization, \cite{pmlr-v119-wang20x,rao2020make}. Another newly documented~\cite{sheikh2020reducing} undesired behavior of TD3  is to have all Critics converge to a same point in parameter space and degenerate into single-Q-network performance. Without further experiments we cannot say if poor performance is due to action clipping, to critic diversity collapse, or both. Considering SAC works a lot better and also uses dual Q-networks like TD3, it seems more likely that clipping and bad exploration are to blame than diversity collapse.  

\paragraph{Quality of deep and shallow controllers}

It is actually hard to find good attitude controllers using RL, probably explaining why papers in this area generally only discuss a single neural net controller: we found only 9 out of about 5000 controllers which complied with our specifications. The very last 5\% performance seems to be very hard to get because of “spikes” we observed, due to spurious correlations in the fully connected neural net controllers we have been considering. We also note that small and rather shallow (two or three hidden layers) networks were observed to be best trained and to be behaving best for attitude control. 

\paragraph{Spurious correlations}

\begin{figure}[htbp]
    \centering
    \includegraphics[width=0.9\linewidth]{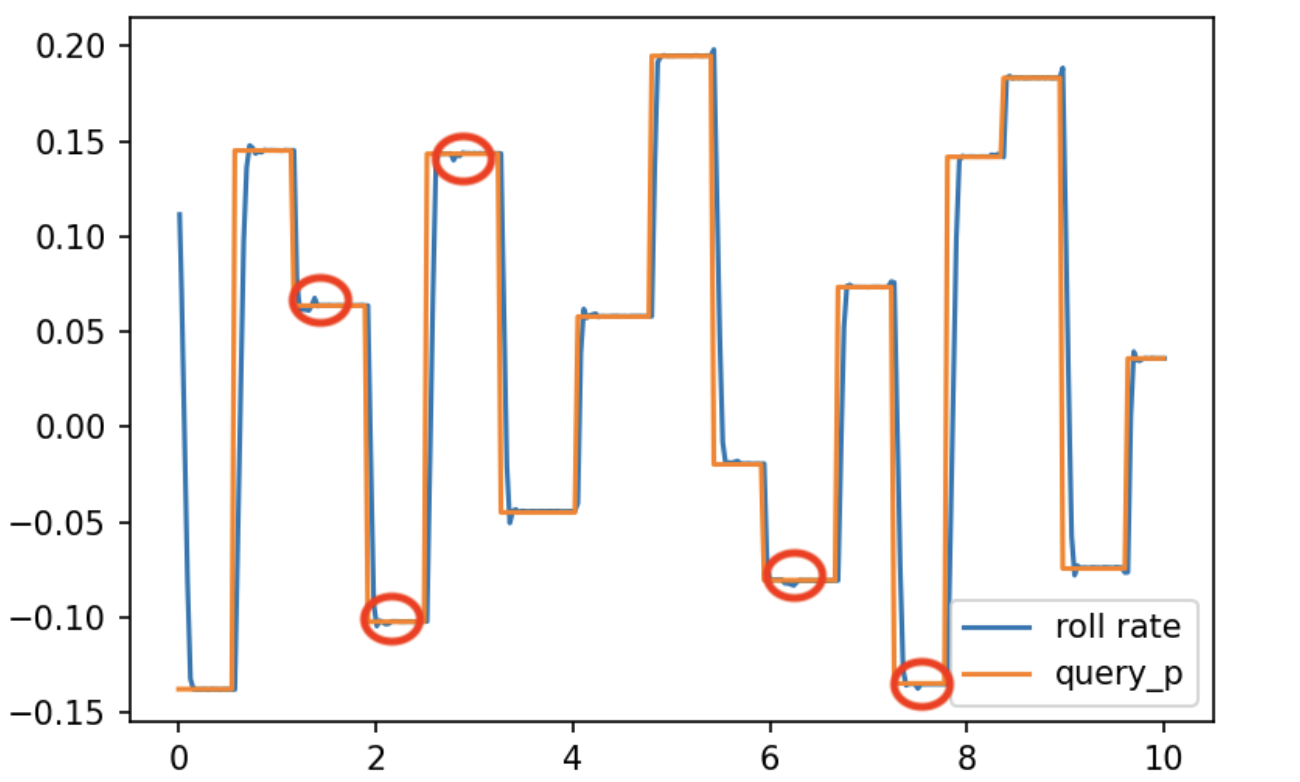}
    \caption{Spikes in roll rate control, with one of our best trained neural nets}
    \label{fig:spikes}
\end{figure}

Even if the STL metrics gives excellent results for some networks, there are still some spikes, as shown with the behavior of one of our best networks on a simple roll rate query in Figure \ref{fig:spikes}, that we identified to be due to spurious correlations between the errors on one axis and the command on another axis. 

\begin{figure}[htbp]
    \centering
    \includegraphics[width=0.9\linewidth]{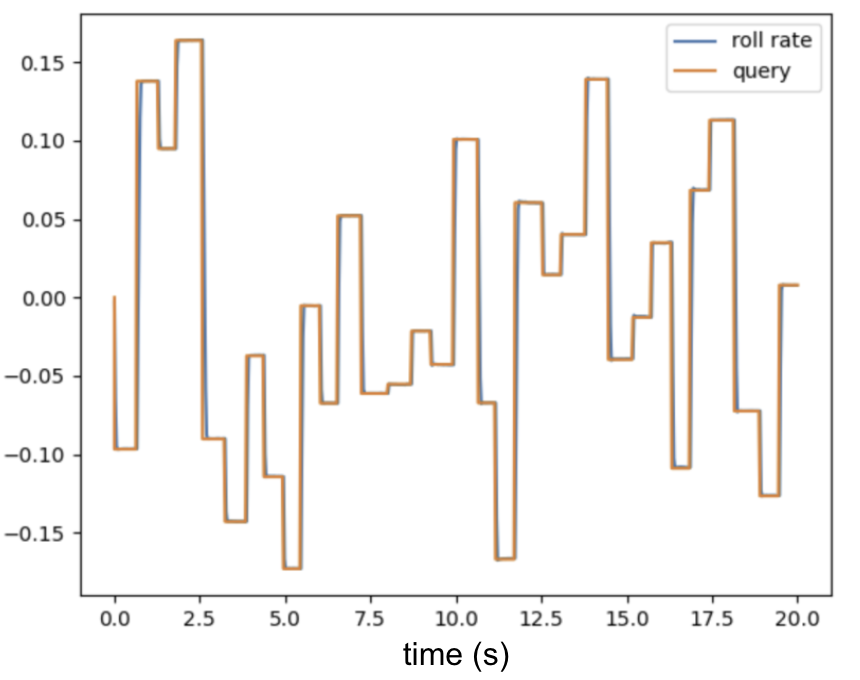}
    \caption{Behaviour of a controller trained on roll only.}
    \label{fig:only-roll-control}
\end{figure}

These spurious correlations can be exposed by 
training a controller on a single axis, here the roll axis, and showing that they indeed do not appear in that case where correlations cannot possibly be made. This new controller was trained  with only the error on the roll rate as input, with the objective of controlling only $cmd_\phi$. During training, we have been controlling $cmd_\psi$ and $cmd_\theta$ by PIDs. We see in Figure \ref{fig:only-roll-control} that the controller on roll rate is now almost perfect, showing no spikes. 

\begin{figure}[htbp]
\centering
\begin{subfigure}[b]{0.45\textwidth} 
    \centering
    \includegraphics[width=0.5\linewidth]{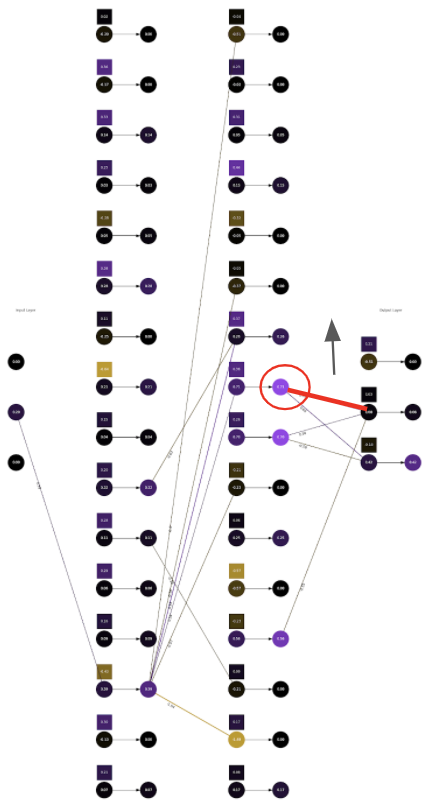}
    \caption{A case of good correlation between the input and the roll rate output}
    \label{fig:good-correlation}
\end{subfigure}
\begin{subfigure}[b]{0.45\textwidth} 
    \centering
    \includegraphics[width=0.5\linewidth]{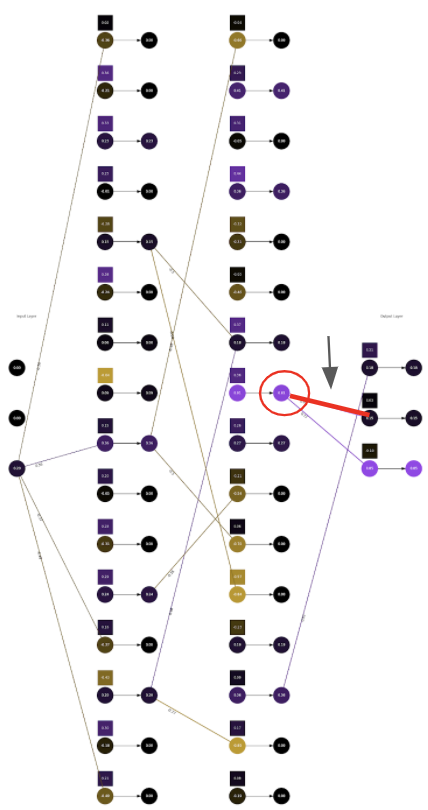}
    \caption{A case of spurious correlation between the input and the roll rate output}
    \label{fig:spurious-correlations}
    \end{subfigure}
    \caption{Two different types of correlations during training}
\end{figure}

Another way to expose the spurious correlations is to examine the connections 
between the neurons of our controller, and in particular to show which ones are above 
a certain threshold, for a given input. 
In Figures \ref{fig:good-correlation} and \ref{fig:spurious-correlations}, we depicted the case of a 16$\times$16 neural net controller for the roll, pitch and yaw rates, with two different inputs. The red arc is the same connection in the two figures, between some neuron of the second hidden layer and the neuron governing the roll rate output. In both cases it has a high value (weight), but in the first case, Figure \ref{fig:good-correlation}, 
it shows a good correlation with the input that is used, while in the other, Figure \ref{fig:spurious-correlations}, it shows a spurious correlation. 
The correlation of Figure \ref{fig:spurious-correlations} is deemed ``good", or correct, since the second input linked to the second axis is connected to the second output (on the same axis - connections between neurons are highlighted in the corresponding figures) and the correlation of Figure \ref{fig:spurious-correlations} is deemed spurious because it happens when the third input linked to the third axis is connected to the second input (an error on the pitch axis should not influence an action on the yaw axis). 

During the training phase, i.e. gradient descent, the weight of the connection will never converge to something sensible enough: when the network sees the first type of input, it will increase the importance of this connection while for the second input, it will reduce the importance of the same connection. 

\paragraph{Training for nominal and non-nominal situations}

We also observed that 
there is some amount of robustness built in neural net controllers, suitably trained in nominal conditions, to certain non-nominal situations. We believe this is due to the fact that the controllers which are trained in the nominal case, are actually trained in many different states that appear in non-nominal situations, for the same neural net inputs (e.g. angular rate errors), by using a very wide distribution of initial states during training. Similar observations on robustness by training from wide initial state distributions were made in \cite{locomote}. 

Finally, training neural net controllers to both nominal and non-nominal situations is not an easy endeavor and should be further studied. The difficulty lies in training on sufficiently many non-nominal data, as well as avoiding over-fitting to non-nominal cases: reward distributions can become multi-modal and expectation maximization could be bad in such cases. 

For instance, when we saturate a motor, we lose a degree of freedom and we can just hope for, for instance, a good control on the roll and pitch axis, at the expense of some degradation for controlling the yaw rate. 
Indeed, it is much harder for the quadcopter to generate a moment on the yaw axis than on the roll or on the pitch axis. 

\begin{figure*}[htbp]
    \centering
    \includegraphics[width=\linewidth]{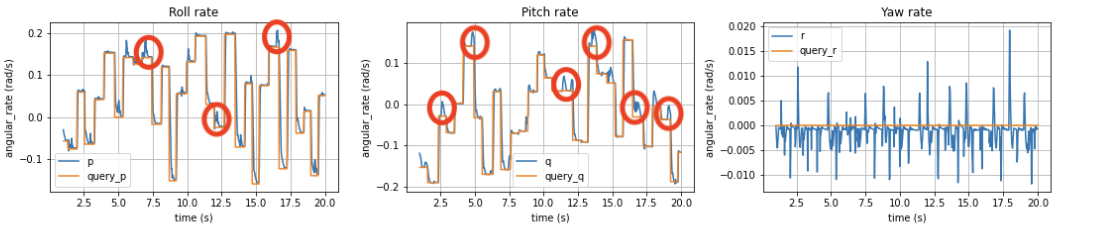}
    \caption{Controller trained in nominal mode and tested with motor saturation}
    \label{fig:trained_nominal_tested_saturation}
\end{figure*}

\begin{figure*}[htbp]
    \centering
    \includegraphics[width=\linewidth]{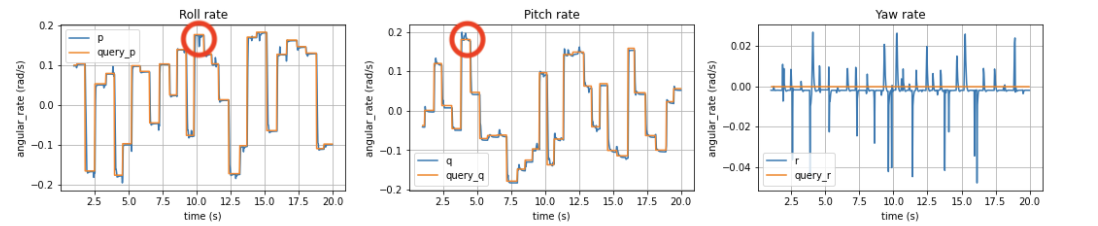}
    \caption{Controller trained with motor saturation and tested with motor saturation}
    \label{fig:trained_saturation_tested_saturation}
\end{figure*}

When the controller has only been trained in nominal mode (i.e. without any saturation), it can stay for a rather long time far from the query, when used in non-nominal mode (here, with one motor saturated to a portion of its power, as explained in Section \ref{sec:motorfailure}), as shown in Figure \ref{fig:trained_nominal_tested_saturation}. 
When the controller is trained in non-nominal mode, it learns to
overcompensate and does not remain far from the query for a long time, see Figure \ref{fig:trained_saturation_tested_saturation}. Indeed, when one motor is saturated, the command on one axis will create a moment on another axis that needs to be compensated. This is what the drone successfully  learns when trained with a saturated motor.

\section{Conclusion}

We have presented a complete study of learned attitude controls for a quadcopter using reinforcement learning. In particular we extend previous results by modeling partial motor failure as well as wind gusts, and generating extensive tests of various network architectures, training algorithms and hyperparameters using a flexible and robust experimental platform. We also present a precise evaluation mechanism based on robust signal temporal logic observers, which allows us to characterize the best options for training attitude controllers.
Results show that learned controllers exhibit high quality over a range of query signals, and are more robust to perturbations than PID controllers.

The immediate next step will be to start
using STL-derived reward signals during training on the most promising
architectures, and try to improve training under non-nominal situations. 

Finally, because we use an explicit ODE model, we can hope to discuss formal reachability properties of the complete controlled system, using or elaborating on approaches such as \cite{sherlock} and \cite{goubault_putot}. 




\bibliography{main}




\end{document}